\documentclass[10pt,twocolumn,letterpaper]{article}

\usepackage{cvpr}      % To produce the REVIEW version

\usepackage{etoc}
\etocdepthtag.toc{mtchapter}
\etocsettagdepth{mtchapter}{subsection}
\etocsettagdepth{mtappendix}{none}
\usepackage{minitoc}
\definecolor{cvprblue}{rgb}{0.21,0.49,0.74}
\usepackage[pagebackref,breaklinks,colorlinks,allcolors=cvprblue]{hyperref}
\usepackage{wrapfig}
\usepackage{pifont}
\usepackage{array}
\usepackage{hyperref}
\usepackage{url}

\usepackage{booktabs,makecell}
\usepackage{wrapfig}

\usepackage{paracol}
\usepackage{amsmath,amssymb,algorithm,algorithmic}
\usepackage{overpic}
\usepackage{adjustbox}
\usepackage{colortbl}
\usepackage{multirow}
\usepackage{booktabs}
\usepackage{tabularx} 
\usepackage{booktabs,multirow,makecell,array,graphicx}
\usepackage{enumitem}
\usepackage{threeparttable}
\usepackage{siunitx}
\usepackage{array}
\usepackage{longtable}
\usepackage{calc}      % for \dimexpr
\usepackage{caption}   % for \captionsetup / \captionof
\usepackage{tikz}
\usetikzlibrary{positioning,arrows.meta}
\usepackage{caption}
\usepackage{subcaption}      % for side-by-side captions
\usepackage{amsmath}

\usepackage{etoc}
\etocdepthtag.toc{mtchapter}
\etocsettagdepth{mtchapter}{subsection}
\etocsettagdepth{mtappendix}{none}
\newcolumntype{L}[1]{>{\raggedright\arraybackslash}p{#1}}

\title{RoboWheel: A Data Engine from Real-World Human Demonstrations for Cross-Embodiment Robotic Learning}

\author{Yuhong Zhang$^{1}$ \thanks{Equal contribution,  \ $^\dag$Corresponding author. Work done by Yuhong Zhang, Zihan Gao during the internship at Synapath Research.} \quad
Zihan Gao$^{1,*}$ \ \hfill
Shengpeng Li$^{2}$ \ \hfill
Ling-Hao Chen$^{1}$ \ \hfill
Kaisheng Liu$^{2}$  \ \hfill 
Runqing Cheng$^{2}$ \\   %\hfill
Xiao Lin$^{2\dag}$ \ \hfill
Junjia Liu$^{2,3}$ \  \hfill
Zhuoheng Li$^{4}$ \  \hfill
Jingyi Feng$^{5}$ \  \hfill
Ziyan He$^{2}$ \  \hfill
Jintian Lin$^{2}$ \\
Zheyan Huang$^{1}$ \  \hfill
Zhifang Liu$^{1}$ \  \hfill
Haoqian Wang$^{1\dag}$ \\
{\small $^1$Tsinghua University \hfill
$^2$Synapath  \hfill
$^3$CUHK  \hfill
$^4$HKU  \hfill
$^5$PolyU} \\
{\texttt{\small \{dsyuhong, gaozihanthu, adreamob, thu.lhchen\}@gmail.com}}\\
Project page: \url{https://zhangyuhong01.github.io/Robowheel}
}

\makeatletter
\apptocmd\@maketitle{{\myfigure{}\par}}{}{}
\makeatother

\newcommand\myfigure{%
\vspace{-2em}
\centering
   \includegraphics[width=\linewidth]{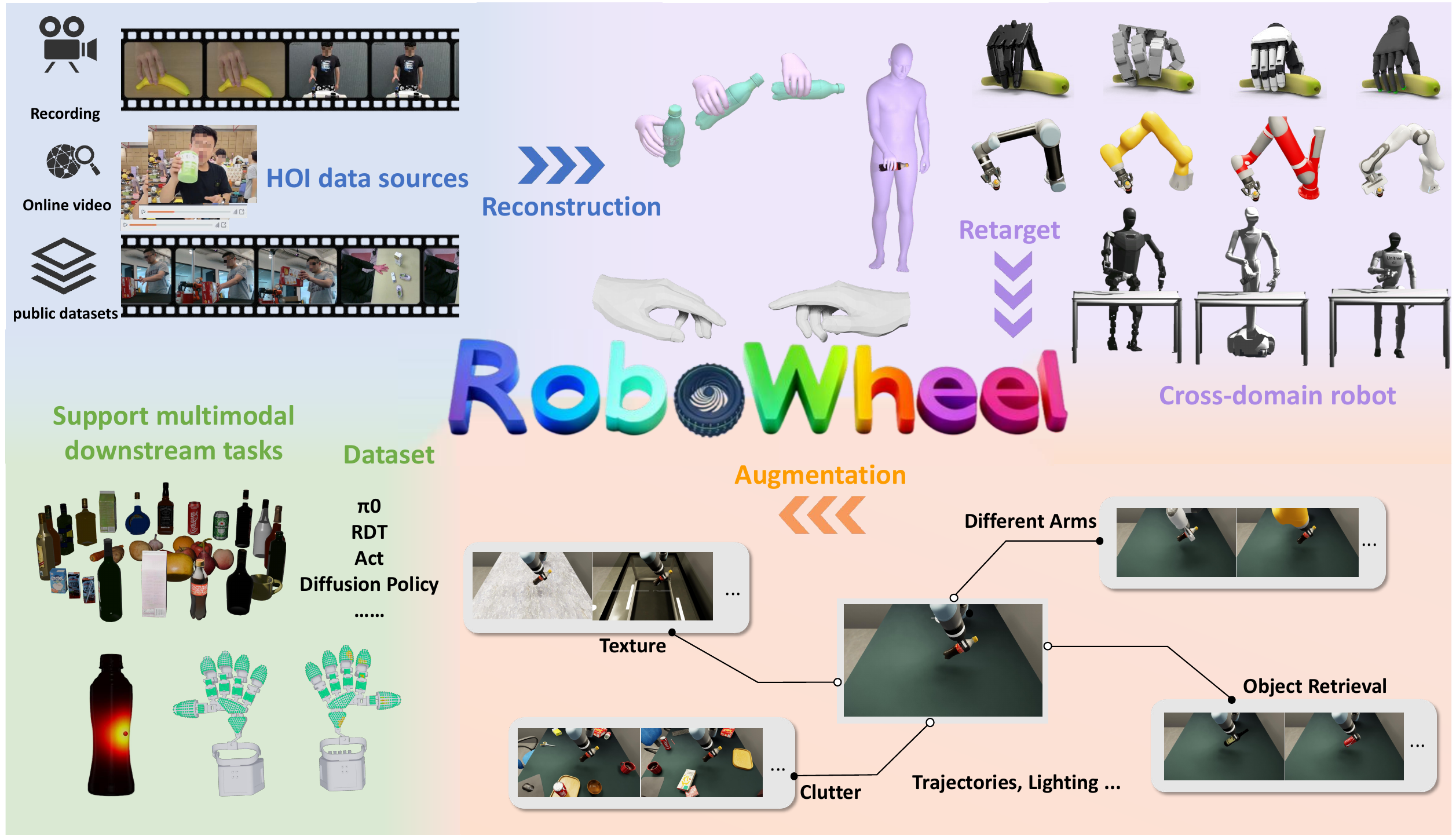}
\vspace{-2em}
\captionof{figure}{The \nbname \ data engine. Our pipeline could process hand-object interaction (HOI) videos from diverse sources (\textit{e.g.}, recordings and public datasets) through high-fidelity reconstruction to recover physically consistent trajectories. The reconstructed motions are retargeted to cross-domain robotic embodiments (\textit{e.g.}, arms, dexterous hands, humanoids), and enhanced with multi-modal data augmentations (\textit{e.g.}, object retrieval, texture, trajectories). This generates a large-scale dataset with multimodal observations, supporting training for various vision-language-action (VLA) and imitation learning models (\textit{e.g.}, ACT, Diffusion Policy).}
\vspace{1em}
\label{fig:teaser}
}
\definecolor{stage1}{RGB}{14,166,233}   
\definecolor{stage2}{RGB}{139,92,246}  
\definecolor{stage3}{RGB}{34,197,94}  

\newcommand{\nbname}{\textit{RoboWheel}}
\newcommand{\nbdata}{\textit{HORA}}

\newcommand{\cmark}{\checkmark} 
\newcommand{\xmark}{\makebox[1.2ex][c]{\rule{1.2ex}{1.2ex}}} % solid square as "✗" 
\begin{document}
\maketitle
\pagestyle{plain}
\thispagestyle{plain}

\begin{abstract}
We introduce \textbf{\nbname}, a data engine that converts human hand–object interaction (HOI) videos into training-ready supervision for cross-morphology robotic learning. From monocular RGB/RGB-D inputs, we perform high-precision HOI reconstruction and enforce physical plausibility via a reinforcement learning (RL) optimizer that refines hand–object relative poses under contact and penetration constraints. The reconstructed, contact-rich trajectories are then retargeted to cross-embodiments, robot arms with simple end-effectors, dexterous hands, and humanoids, yielding executable actions and rollouts. To scale coverage, we build a simulation-augmented framework on Isaac Sim with diverse domain randomization (embodiments, trajectories, object retrieval, background textures, hand motion mirroring), which enriches the distributions of trajectories and observations while preserving spatial relationships and physical plausibility. The entire data pipeline forms an end-to-end pipeline from video → reconstruction → retargeting → augmentation → data acquisition. 
We validate the data on mainstream vision–language–action (VLA) and imitation learning architectures, demonstrating that trajectories produced by our pipeline are as stable as those from teleoperation and yield comparable continual performance gains. To our knowledge, this provides the first quantitative evidence that HOI modalities can serve as effective supervision for robotic learning. Compared with teleoperation, \nbname \ is lightweight: a single monocular RGB(D) camera is sufficient to extract a universal, embodiment-agnostic motion representation that could be flexibly retargeted across embodiments. We further assemble a large-scale multimodal dataset combining multi-camera captures, monocular videos, and public HOI corpora for training and evaluating embodied models.

\end{abstract}

\vspace{-1.5em}
\section{Introduction}
\vspace{-0.3em}
\label{sec:intro}
% Embodied agents learn most effectively when supervision reflects how humans actually interact with the physical world. 
% However, obtaining contact-rich, robot-usable supervision at scale remains notoriously difficult. Existing pipelines typically rely on prelabeled, curated human video datasets or studio motion capture, which limits coverage, diversity, and transfer across different embodiments and tasks. Notably, the Internet contains an immense reservoir of hand–object interaction (HOI) videos (hand-only or upper-body) that include rich manipulation strategies, but these signals are rarely converted into training-ready data for robots due to reconstruction noise, physical implausibility, and embodiment mismatching.
  % \vspace{-0.3em}
Embodied agents learn most effectively when their supervision matches how humans interact with the world. However, obtaining specific robot-usable supervision at scale is challenging. Existing data collection pipelines mainly rely on teleoperation or studio motion capture, which demands specialized hardware and careful curation, limiting the diversity of action behaviors and their transferability across robot embodiments and tasks. At the same time, multiple data sources provide vast amounts of hand–object interaction (HOI) information containing rich and real-world manipulation strategies, yet these signals are rarely converted into training-ready data for robots due to reconstruction noise, physical implausibility, and embodiment mismatch.

Recent advances in human- and object-centric perception suggest an opportunity to close this gap. Methods based on SMPL-H/MANO ~\cite{zhang2025hawor, zhang2025humanmm, yin2024whac}
% hand/body models 
and 6D object pose or mesh tracking ~\cite{zhang2024omni6dpose, wen2025foundationstereo, fpose} can recover detailed geometry and motion from monocular RGB/RGB-D inputs. Nevertheless, contact estimates are often inconsistent, hand–object interpenetrations arise under occlusion, trajectories lack temporal smoothness, and the recovered motions generally violate kinematic and dynamic constraints. 
%This disconnect highlights a persistent gap between what can be reconstructed from HOI videos and what a robot can reliably execute or learn from.

Bridging this gap requires a practical, scalable processing pipeline that satisfies three key requirements: (i) enabling large-scale, continuous acquisition of robot–object interaction trajectories in real-world operational spaces while enforcing physical plausibility; (ii) supporting flexible retargeting of these trajectories to diverse robot embodiments—even across domains—while preserving interaction semantics; and (iii) maintaining scalability by enabling effective composition of data augmentation strategies. Existing paradigms for robot data collection—teleoperation and simulated data generation—fail to meet these requirements jointly: teleoperation is costly and hardware-specific, whereas purely synthetic simulation data often does not reflect real-world perceptual and contact distributions.

We address these challenges with \nbname, a data engine that transforms real-world hand–object demonstrations into supervision for cross-embodiment robotic learning. Starting from monocular RGB(D) HOI videos, \nbname \ integrates state-of-the-art hand, body, and object motion estimators into a unified reconstruction framework. A multi-stage physical plausibility optimization then refines the trajectories: signed distance function (SDF)-based penalties discourage interpenetration and encourage contact, and a residual reinforcement learning (RL) policy further adjusts hand–object relative poses under reachability and stability priors. The resulting trajectories are mapped into a canonical action space and retargeted to heterogeneous robot morphologies—including 6/7-DoF arms with parallel grippers, dexterous hands, and humanoids, and output executable control sequences in both operational and joint spaces.

% To scale coverage while preserving interaction semantics, 
% % \nbname \  further builds a simulation-augmented data flywheel. 
% retargeted trajectories are replayed in simulation across multiple arms using GPU-accelerated inverse kinematics.
% Within this environment,we apply domain randomization and augmentation, including embodiment variations, trajectory variations, object replacement, background texture randomization, hand mirroring, and more.
To scale coverage while preserving interaction semantics, we replay the retargeted trajectories in simulation across multiple robotic arms using GPU-accelerated parallel inverse kinematics. Within this environment, we apply domain randomization and diverse augmentations for observation and robotic action, including embodiment variations, trajectory distribution enrichment, object retrieval and replacement, background texture randomization, cluttered table configurations, hand mirroring, and more.
% —varying embodiments, mirroring hand motions, trajectories, replacing objects with geometrically and semantically compatible assets, and randomizing background appearance. 
% These procedures enrich the distributions of observations and trajectories while maintaining physically plausible hand–object relationships. We validate the resulting data on a range of VLA and imitation learning (IL) models, and show that policies trained on \nbname-generated trajectories can match the performance of policies trained on teleoperation data.
%while pretraining on \nbname data significantly improves robustness under distribution shifts.

On top of this engine, we construct \nbdata \ (\textbf{H}and–\textbf{O}bject to \textbf{R}obot \textbf{A}ction dataset), a large-scale multimodal dataset from three sources: a custom multi-view motion capture system with tactile-sensor gloves, an RGB(D) HOI recording setup, and multiple public HOI datasets. 
% \nbname \  converts these heterogeneous inputs into a unified representation that jointly captures HOI and robot-execution modalities, including hand/body parameters, 6D object motion, object assets, contact annotations, multi-view and wrist-view observations, and, for the mocap subset, dense tactile maps. 
As a multimodal dataset, \nbdata \  combines the data modalities of HOI corpora and embodied robot datasets and \nbdata \  supports both HOI-related analysis and downstream robotic learning.

In summary, our key contributions as follows.

\begin{itemize}[leftmargin=*]
  \item \textbf{Physically plausible HOI reconstruction and cross-domain retargeting.}
  % A contact-consistent HOI reconstruction framework from monocular RGB/RGB-D that combines SOTA hand/whole-body/object motion estimation with multi-stage physical optimization, unlocking HOI videos as a scalable supervision source for robotic learning.
    A monocular RGB/RGB-D HOI reconstruction framework that integrates state-of-the-art hand, body, and object motion estimation with physics-based optimization. The framework supports flexible cross-embodiment retargeting, outputs executable motion trajectories, and provides scalable supervision signals for downstream learning.
% \item \textbf{Cross-embodiment robust retargeting.}
  % A general retargeting module that produces robot-usable supervision across diverse embodiments (robot arms, dexterous hands, humanoids), with executable trajectories in both operational (end-effector) and joint spaces.
  \item \textbf{Simulation-augmented data flywheel.}
  % HOI-conditioned augmentation and domain randomization in Isaac Sim (hand mirroring, embodiment variants, object replacement, background changes), with validation on mainstream VLA and imitation-learning settings.
  %A diverse augmentation and domain randomization based on Isaac Sim (embodiment variants, object replacement, background variation, hands mirroring, etc.) conditioned on HOI. This data flywheel is validated on mainstream VLA and imitation-learning settings to enhance robustness and scalability in robotic learning.
  We implement rich augmentation and domain randomization in simulation driven by HOI data. This data flywheel is validated on mainstream VLA and imitation learning models, and we conduct a quantitative evaluation of the quality and effectiveness of both the HOI transformation and the augmentation strategy.
  \item \textbf{Large-scale multimodal dataset.}
  A multimodal dataset of over 150,000 sequences is constructed from multi-view motion capture, public HOI datasets, and recorded videos. It includes embodiment trajectories, observations, assets, HOI annotations, and task descriptions, with a mocap-based subset also providing tactile signals, offering a rich and scalable resource for robotic learning and downstream HOI-related tasks.

% Thousands of high-precision sequences aggregated from an in-house multi-camera/mocap pipeline, public HOI datasets, and curated online videos, including trajectories, contacts, multi-view obserbation. This dataset is validated across multiple model types, ensuring versatility and effectiveness in real-world applications.
\end{itemize}
\vspace{-0.5em}
\section{Related Work}
\vspace{-0.5em}
\label{sec:related_work}

\paragraph{HOI datasets and monocular reconstruction.}
High-precision HOI annotations remain costly, as most public 3D HOI datasets rely on multiview rigs or motion capture (MoCap) systems for accurate hand-object geometry~\citep{Chao_2021_CVPR,hampali2020honnotate,hampali2021ho3dv3,taheri2020grab,wang2024ho}. Large egocentric video corpora like ~\cite{grauman2024ego} use head-mounted cameras to avoid MoCap but lack frame-accurate 3D HOI geometry for reconstruction~\cite{grauman2022ego4d}. Recent whole-body motion datasets such as~\cite{zhang2025motionxpp} scale to millions of SMPL-X frames but are not dedicated HOI datasets and offer limited hand-object contact supervision. On the algorithmic side, ~\cite{chen2025hort} reconstructs objects by fusing pixel-aligned features with 3D hand geometry in a transformer-based coarse-to-fine point cloud decoder, yielding dense object geometry with high frame fidelity, while ~\cite{fan2024hold} jointly reconstructs articulated hands and objects using compositional SDF and contact constraints. These methods, however, are generally limited to single-frame or in-contact scenarios and struggle with approach/withdrawal phases, generalizability, occlusion, low video resolution, and varying hand movement speeds. Recently, more generalizable approaches~\citep{prakash2023learning, yang2023boosting, qu2023novel} have used data-driven priors; for instance, \cite{yang2023boosting} introduces diffusion-guided, per-video optimization to enhance robustness under occlusion, albeit at the cost of heavier computation and the need for short clips.

\vspace{-1.3em}

\paragraph{Robotic Learning from Human Demonstration}
Early work in vision-based programming by demonstration mapped human hand poses to robot grasps directly from images, establishing a pipeline from grasp recognition to example-based robot execution~\cite{kjellstrom2008visual}. More recent systems leverage richer HOI signals: ~\cite{zhou2025you} extract binocular hand-motion cues from human videos, compress trajectories into keyframes with coordination masks, augment demonstrations geometrically, and train a bimanual diffusion policy that executes long-horizon dual-arm tasks and generalizes across scenes. 
Complementarily, ~\cite{gatgrasp2025} treat human gestures as structured priors, retrieving grasp affordances from HOI memories and transferring them to novel objects, yielding robust performance in single-object and cluttered settings.
At scale, cross-embodiment corpora and models (Open X-Embodiment/RT-X) demonstrate positive transfer across heterogeneous robots, motivating retargetable supervision from human interactions~\citep{rtx2023}.
Specialized transfer frameworks extend this idea to dexterous bimanual manipulation via residual learning~\citep{maniptrans2025}, while whole-body humanoid control benefits from motion-tracking pipelines distilled into guided diffusion policies that enable versatile downstream behaviors~\citep{beyondmimic2025}. Together, these threads indicate a viable route from human HOI video to robot-usable policies via reconstruction, retargeting, and augmentation.

% % \paragraph{Robotic Learning from Human Demonstration}
% Early work in vision-based programming by demonstration directly mapped human hand poses to robot grasps from images, establishing a grasp recognition-to-execution pipeline~\citep{kjellstrom2008visual}. More recent approaches, such as~\cite{zhou2025you}, extract binocular hand-motion cues, compress trajectories into keyframes, and train a bimanual diffusion policy to generalize long-horizon tasks across scenes. \cite{gatgrasp2025} treat human gestures as structured priors, retrieving grasp affordances from HOI memories and applying them to novel objects, improving performance in both single-object and cluttered environments. At scale, cross-embodiment models (Open X-Embodiment/RT-X) show positive transfer across heterogeneous robots, enabling retargetable supervision from human interactions~\citep{rtx2023}. Specialized transfer frameworks use residual learning for dexterous bimanual manipulation~\citep{maniptrans2025}, while motion-tracking pipelines distilled into guided diffusion policies improve whole-body humanoid control~\citep{beyondmimic2025}. These approaches collectively pave the way from human HOI videos to robot-usable policies through reconstruction, retargeting, and data augmentation.

\vspace{-1.3em}

\paragraph{Embodied models and scalable data for generalist manipulation.}
Generalist \textit{vision--language--action} policies pretrained on large video and robot corpora to enable instruction following and out-of-distribution generalization across tasks and embodiments~\citep{brohan2022rt1,zitkovich2023rt,kim2024openvla,pi05_2025}. In parallel, imitation- and diffusion-based visuomotor learning emphasize stable training and multimodal action distributions, from classic action-diffusion policies to large diffusion foundation models that scale to bimanual control~\cite{chi2023diffusionpolicy,liu2024rdt1b}. To reduce data and hardware barriers, low-cost bimanual teleoperation systems provide dense demonstrations for fine-grained skills~\citep{zhao2023aloha}, while object-/pose-centric representations and semantic flows improve cross-object generalization and pose awareness~\cite{chen2025g3flow}. At the dataset/benchmark layer, dual-arm generators and domain-randomized platforms supply scalable supervision with unified evaluation~\citep{mu2025robotwin,chen2025robotwin2}; open-instruction rearrangement benchmarks probe 6-DoF reasoning under language guidance~\citep{ding2024open6dor}; and video-driven pipelines synthesize long-horizon tasks directly from Internet videos~\citep{ye2025video2policy}. Recent work on task-centric \emph{latent actions} further mitigates embodiment mismatch by learning instruction-conditioned action spaces transferable across robots~\citep{bu2025univla}. 

% Due to page limitations, we leave \textit{Robotic Learning from Human Demonstration} discussion in~\cref{relatedwork}.

% Together, these lines suggest a complementary recipe—\emph{VLA pretraining + robust visuomotor learning + scalable, embodiment-aware data}—and motivate treating HOI video as a first-class supervision source via reconstruction, retargeting, and augmentation.
% \input{sec/problem_chanllenge}
\vspace{-0.3em}
\section{Method}
\vspace{-0.2em}
\label{sec:method}

\subsection{System Overview}
\begin{figure*}[h]
    \centering
    \vspace{-0.5em}
    \begin{overpic}[width=\linewidth]{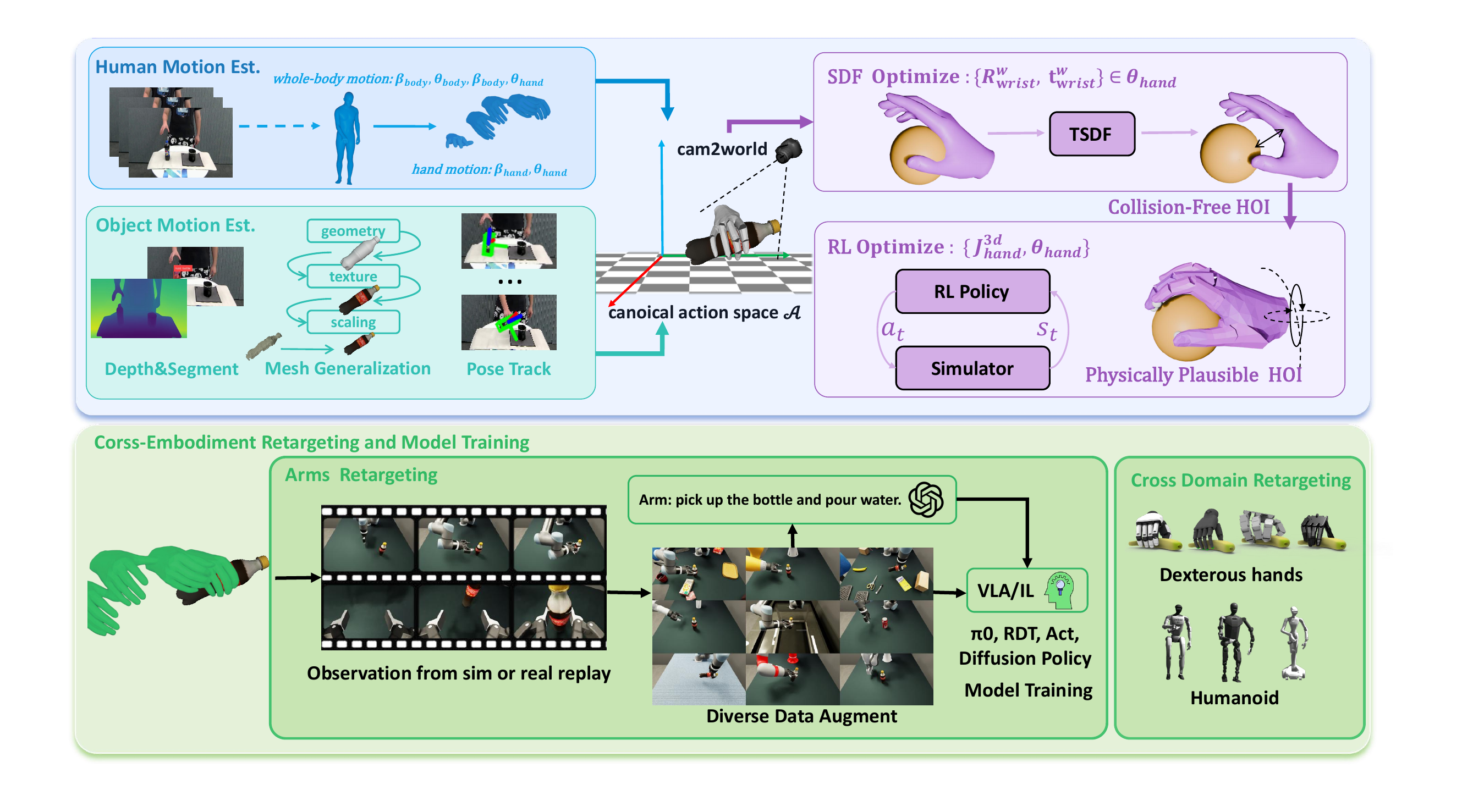}
    \end{overpic}
    \vspace{-2em}
    \caption{Given monocular RGB(-D) input, we first estimate the motion of the hand or wholebody and the manipulated object. We then perform a joint optimization, guided by TSDF and reinforcement learning, to improve physical plausibility and ensure robotic reachability. The resulting trajectories are retargeted to heterogeneous embodiments—including arms, dexterous hands, and humanoids. Finally, domain randomization for both observation and trajectories in Isaac Sim is applied to enrich observational diversity for robotic arms, and the generated embodied data is validated across both VLA and IL policy benchmarks.}
    \vspace{-1.3em}
    \label{fig:pipeline}
\end{figure*}

\vspace{-0.2em}
We build a systematic pipeline covering in-the-wild hand-object interaction(HOI) videos into robot-usable supervision data. Our pipeline overview is illustrated in Fig.~\ref{fig:pipeline}.

\vspace{-0.5em}
\subsection{HOI Reconstruction from RGB(D) videos}
\label{sec:optim goal}

\vspace{-0.5em}
\paragraph{Problem setup.} Given video frames $\{I_t\}_{t=1}^T$, our goal is to recover metrically consistent trajectories together with parametric representation of the hand and mesh of the manipulated object in the same world coordinate ,while (i) preventing hand–object interpenetration and (ii) enforcing physically plausible, temporally stable contact.Concretely, the state of the hand pose at time $t$ is, 
\vspace{-0.5em}
\begin{equation}
    \mathbf{h}_t \;=\; (\mathbf{\theta}_h(t),\, \mathbf{R}_h^w(t),\, \mathbf{t}_h^w(t)),
    \label{eq:hand-state}
\vspace{-0.5em}
\end{equation}
where $\mathbf{\theta}_h(t)$ is the hand pose, $\mathbf{R}_h^w(t)$ and $\mathbf{t}_h^w(t)$ are the global transform and wrist of hands in the world coordinate. The object state is the rigid 6D pose tied to its (scale-resolved) geometry,
% \begin{equation}
$\mathbf{p}_t \;=\; T_o^w(t) \in \mathtt{SE}(3)$,
% \label{eq:object-state}
% \end{equation}
defines the location and rotation of the object. 

  \vspace{-1em}

% Let $K$ denote the intrinsics of the camera and let $T_c^w=(R_{wc},\,t_{wc})\in \mathrm{SE}(3)$ be the (time-invariant) camera-to-world transform estimated by ~\cite{droid}. We use perspective projection as $\Pi\!\left(K, R, t; X\right)$ and use homogeneous composition for transforms. 
%Under this parameterization, the cam$\!\to\!$world lifting introduced below maps per-frame camera states to world states, yielding $\{\theta_h(t), R_h^w(t), t_h^w(t)\}$ for the hand and $T_o^w(t)$ for the object; the explicit compositions remain as in Eqs.~\eqref{eq:hand-world} and \eqref{eq:obj-world}.

\paragraph{Human and hand motion recovery.}
Our method initially determines whether a clip implies \emph{hand-only} or \emph{whole-body} motion. For the \emph{hand-only} case, we estimate $\mathbf{h}_t$ per frame using~\cite{hamer}. Otherwise, we estimate the SMPL-H parameters via~\cite{zhang2025motionxpp} and directly produce the world coordinate body pose $\theta_b(t)$ and the shape $\beta_b$, equivalently extracting the hand state $\mathbf{h}_t$.

  \vspace{-1em}

\paragraph{Object reconstruction and pose estimation.}
We ground the manipulated object, obtaining the per-frame mask $m_t$ and depth $D_t$ (predicted by ~\cite{unidepth} or RGB-D) in the video. 
Conditioned on semantic cues, we use a multiview 3D generator $\mathcal{G}$~\citep{hunyuan3d2} to produce an unscaled textured mesh $\hat{M}_o$. 
Then we recover the metric scale of the manipulated object by back-projecting the depth map inside the mask to a point set
$\mathcal{P}_t=\{X_c(p)=D_t(p)\,K^{-1}\tilde{p}\mid p\in m_t\}$, aggregating as $\mathcal{P}=\bigcup_t\mathcal{P}_t$.
We obtain the rescaled object mesh $M_o$ via $M_o \;=\; s_o\,\hat{M}_o,$ where $s_o $ denotes the ratio between the diagonals of the bounding boxes of $\mathcal{P}$ and $\hat{M}_o$. 
% We determine the scaling factor for $\hat{M}_o$ from the ratio of the AABB diagonal lengths of \mathcal{P} and $\hat{M}_o$.
% Letting $\mathrm{diag}(\cdot)$ denote the diagonal of the axis-aligned bounding box $\mathrm{AABB}(\cdot)$, we set $M_o$ as the estimated rescaled object,
% \vspace{-0.5em}
% \begin{equation}
%     M_o \;=\; s_o\,\hat{M}_o,
%     \qquad 
%     s_o 
%     \;=\;
%     {\left\|\mathrm{diag}\!\left(\mathrm{AABB}(\mathcal{P})\right)\right\|_2}
%          /{\|\mathrm{diag}\!\left(\mathrm{AABB}(\hat{M}_o)\right)\|_2}.
%     \label{eq:metric-scale}
%     \vspace{-0.7em}
% \end{equation} 
% \vspace{-0.3em}
% where $s_o$ is the estimated scale factor. 
With $(M_o, M_t, D_t)$, a correspondence-driven tracker $\mathcal{F}(\cdot)$~\citep{fpose} estimates the pose stream of the camera frame object $T_o^c(t) $.
% \begin{equation}
%     T_o^c(t) \;=\; \mathcal{F}\!\left(I_t,\,D_t,\,M_t,\,M_o\right).
%     \label{eq:fp}
% \end{equation}

%\paragraph{Project to a unified action space.}
To eliminate viewpoint-dependent inconsistencies in real-world HOI videos, we first estimate the camera intrinsics \( K \) and the camera-to-world transformation \( T_c^w = (R_{wc}, t_{wc}) \) using ~\cite{droid}. This allows us to transform all reconstructed hand-object interactions to the world coordinate system. We then align the resulting trajectories to a \textbf{canonical action space \( \mathcal{A} \)} by constructing a reference frame based on body joint positions, ensuring consistency across heterogeneous sources.  For detailed transformation steps, please refer to Appendix.

\vspace{-0.7em}

\vspace{-0.5em}
\paragraph{Optimization for physical plausibility.}
%\paragraph{Phase (A): Collision-free optimization}
Let $\phi_o(\mathbf{x};t)$ be a watertight object Truncated Signed Distance Function (TSDF, positive outside), and hand vertices $V_{h}$.
% The optimization pipeline is as follows.
First, we optimize the hand parameter $\mathbf{t}_{h}^{w}$ to avoid penetration between object and hand-palm by minimizing $\phi_o^2(V_h^{palm},t)$.
% \vspace{-0.5em}
% \[
% \mathcal{L}_{\mathrm{h-pen}} = \sum_{t}\sum_{i} \left[\max \left(0, -\phi_{o}(v_{h}^{i};t)\right)\right]^{2}, \quad i \in \left\{ i \mid v_{h}^{i} \in V_{h}^{\mathrm{palm}}(\mathbf{t}_{h}) \right\}.\vspace{-0.5em}
% \]
% As indicated in the upper-right corner of Figure X. Then, we optimize the hand parameter $\mathbf{R}_{h}$, $\mathbf{t}_{h}$, and $\boldsymbol{\theta}_{h}$ to avoid penetration between the hand and the object and achieve reasonable grasping pose, the concrete losses can be viewed in supplementary materials.

%As shown in the upper-right corner of Fig.~\ref{fig:pipeline}, we then optimize the hand parameters ${\mathbf{R_{wrist}}, \mathbf{t_{wrist}}} \in \boldsymbol{\theta}_{h}$,  to avoid hand–object interpenetration and to obtain a reasonable grasping pose. The detailed loss terms are provided in the supplementary material.
As shown in the upper-right corner of Fig.~\ref{fig:pipeline}, we then optimize the hand parameters ${\mathbf{R_{wrist}^{w}}, \mathbf{t_{wrist}^{w}}} \in \boldsymbol{\theta}_{h}$,  to avoid hand–object interpenetration. Detailed collision optimization strategy is described in the supplementary material.
With collision-free HOI initialization, we introduce a residual RL policy~\citep{maniptrans} to obtain physically plausible hand–object poses while ensuring reachability on the robot. Specifically, a residual learning strategy is applied to refine the trajectories of both the hand and the object. The RL policy then encourages accurate tracking of these refined trajectories while promoting physically consistent contact once the hand–object distance drops below a threshold. The resulting state after RL is denoted as \(s_t = \big(h_t,\,p_t,\,\dot{h}_t,\,\dot{p}_t,\,\mathcal{C}_t\big)\). Here,\(h_t,\,p_t\) denote the hand poses and object poses, respectively, along with their corresponding velocities and \(\mathcal{C}_t\) denotes the contact force.Reward function \( r_t \) is defined as follows:
% \[
% R_t = R_{\text{hand}} + R_{\text{object}} + R_{\text{contact}},
% \]

% \textbf{MDP.} State \(s_t = \big(q_h(t),\,T_h(t),\,T_o(t),\,\dot{q}_h(t),\,\mathcal{C}_t\big)\),
% action \(a_t=\Delta q_h(t)\) (or \(\Delta T_h\)), transition by physics.
% The objective maximizes \( \mathbb{E}_\pi \big[ \sum_t \gamma^t r_t \big] \) with

% \begin{align}
% r_t = 
% \underbrace{\lambda_{\text{geo}} R_{\text{geo}} \left( - \lVert \Delta T_o(t) \rVert_{\mathrm{SE}(3)} - \lVert \Delta T_h(t) \rVert_{\mathrm{SE}(3)} \right)}_{\text{geometric reward}} \\
% + \underbrace{\lambda_{\text{dyn}} R_{\text{dyn}} \left( - \lVert \Delta \dot{q}_h(t) \rVert_{\mathrm{SE}(3)} - \lVert \Delta \dot{q}_o(t) \rVert_{\mathrm{SE}(3)} \right)}_{\text{kinematic reward}} \\
% + \underbrace{\lambda_{\text{con}} R_{\text{con}} \left( \mathcal{C}_t \right)}_{\text{contact reward}}.
% \end{align}
% \begin{align}
% r_t = 
% \underbrace{\lambda_{\text{geo}} R_{\text{geo}} \left( - \lVert \Delta T_o(t) \rVert- \lVert \Delta T_h(t) \rVert\right)}_{\text{geometric reward}} 
% + \underbrace{\lambda_{\text{dyn}} R_{\text{dyn}} \left( - \lVert \Delta \dot{q}_h(t) \rVert- \lVert \Delta \dot{q}_o(t) \rVert \right)}_{\text{kinematic reward}} 
% + \underbrace{\lambda_{\text{con}} R_{\text{con}} \left( \mathcal{C}_t \right)}_{\text{contact reward}}.
% \end{align}
% where \Delta  represents the error between the simulated state and the target state.
\vspace{-1em}
% \begin{equation*}
% r_t = 
% \underbrace{\lambda_{\text{geo}} \Phi_{\text{geo}} \left( - \lVert \Delta h_t \rVert - \lVert \Delta p_t \rVert \right)}_{\text{geometric reward}} 
% + \underbrace{\lambda_{\text{dyn}} \Phi_{\text{dyn}} \left( - \lVert \Delta \dot{h}_t \rVert - \lVert \Delta \dot{p}_t \rVert \right)}_{\text{kinematic reward}} 
% + \underbrace{\lambda_{\text{con}} \Phi_{\text{con}} \left( \mathcal{C}_t \right)}_{\text{contact reward}},
% \vspace{-0.5em}
% \end{equation*}

  \vspace{-0.5em}
\begin{equation}
\begin{aligned}
r_t ={}
 \lambda_{\text{geo}} \Phi_{\text{geo}} \left( \lVert \Delta h_t \rVert + \lVert \Delta p_t \rVert \right) \\
 \lambda_{\text{dyn}} \Phi_{\text{dyn}} \left(  \lVert \Delta \dot{h}_t \rVert + \lVert \Delta \dot{p}_t \rVert \right) 
&+ \lambda_{\text{con}} \Phi_{\text{con}} \left( \mathcal{C}_t \right),
\end{aligned}
  \vspace{-0.5em}
\end{equation}
where $\Phi$ denotes the reward function and $\Delta$ denotes the error between resulting states \(s_t\) and target states.

% where \( \psi_{\text{limits}} \) rewards the margin to the joint limits and \( d_{\mathrm{SE}(3)} \) is the geodesic distance on \(\mathrm{SE}(3)\).
% A residual policy is used at control time:
% \( a_t = a_t^{\text{IK}} + \pi_\theta(s_t) \), following ManipTrans-style residual learning.

% \begin{align}
% r_t &= 
% \underbrace{w_{\text{stab}}\,(-\lVert \Delta T_o(t)\rVert_{\mathrm{SE}(3)})}_{\text{object stability}}
% + \underbrace{w_{\text{reach}}\,\psi_{\text{limits}}(q_h(t))}_{\text{reachability}} \\
% \nonumber
% \end{align}

% \vspace{-0.3em}

\subsection{Cross-Embodiment Retargeting}
\label{section:crosser}
\vspace{-0.3em}

%The joint HOI reconstruction by Sec.\ref{sec:optim goal} physically plausible trajectories $\{h_t,p_t\}_{t=1}^T$ with stable hand--object contacts.
%Given these high-quality sequences even from HOI existing public datasets, we seek to retarget them to heterogeneous robot embodiments---industrial arms, dexterous hands, and humanoids.
Based on the physically plausible joint HOI reconstruction in Section \ref{sec:optim goal}, we obtained physically plausible trajectories $\{h_t,p_t\}_{t=1}^T$ and ensured stable hand-object contacts. We aim to retarget these to heterogeneous robot embodiments—industrial arms, dexterous hands, and humanoids.

\vspace{-0.6em}

\begin{figure}[t]
    \centering
\vspace{-0.4em}
    \begin{overpic}[width=0.9\linewidth]{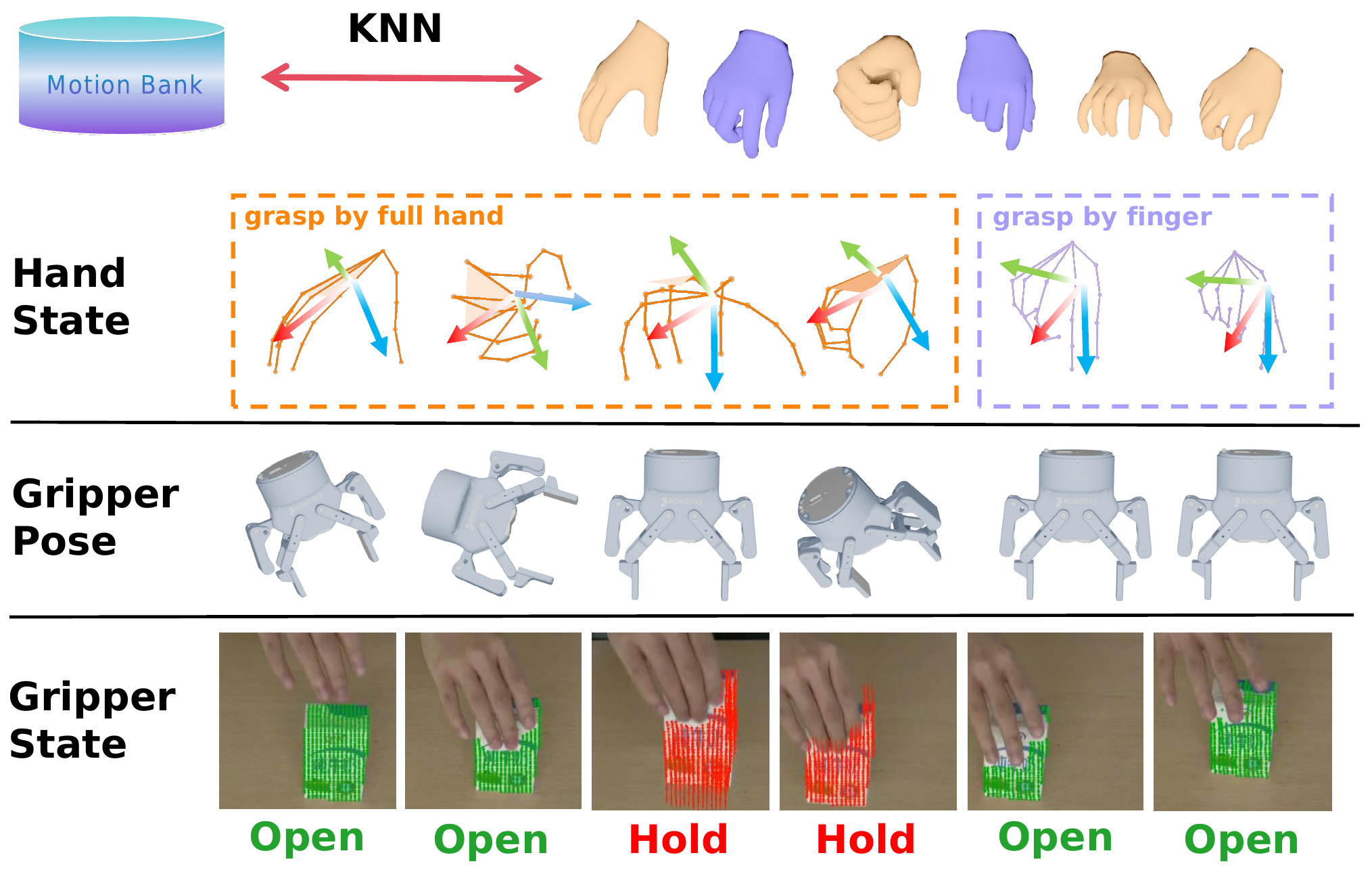}
    \end{overpic}
    \captionsetup{font=footnotesize}
    \vspace{-1.0em}
    \caption{Map the hand’s joints to the gripper’s end-effector pose, including the corresponding mapping of the gripper’s opening and closing state.}
    \vspace{-1.3em}
    \label{fig:retargeting_gripper}
\end{figure}

\vspace{-0.3em}

\paragraph{Robot arms.}
Given accurate 3D hand joints, we retarget hand poses into executable end-effector poses \(\{T_g(t), g(t)\}_{t=1}^T\) for a parallel-jaw gripper (\cref{fig:retargeting_gripper}). Inspired by~\cite{retarget2008}, we implement two complementary orientation constructions depending on whether the \emph{whole hand} (palm-involved) or \emph{only finger tips} dominate the contact geometry. We use a kNN classifier to determine the gesture category.\emph{Whole-hand} retargeting builds a stable palm frame from MCP joints to suppress fingertip jitter; \emph{finger-only} mapping aligns to a hand-intrinsic frame and uses the index--thumb chord to define the gripper axis. For the detailed retargeting algorithm, please refer to the Appendix.To assess the state of the gripper, we employ CoTracker \cite{karaev2024cotracker3} to track the motion trajectories of key points on the manipulated object. The gripper state is determined based on the displacement of these key points.If these points remain stationary, the gripper is classified as open; otherwise, it is classified as closed. By focusing on keypoints rather than the object’s mask, this approach achieves robustness to the visual ambiguity induced by severe occlusions during manipulation.
We subsequently conducted real-robot experiments on a UR5 robotic arm equipped with a parallel gripper to validate the effectiveness of this retargeting method.
% (Kjellstr\"om et al. 2008: classify grasp, then parameterize a predefined strategy by relative hand--object pose; imitate human approach vector.) recognition-based mappingschemes that select a predefined robot grasp and parameterize it by the hand--object relative pose.

% Given accurate 3D hand joints, we retarget hand poses to a parallel-jaw gripper pose. 
% We consider two operation modes: (i) \emph{Whole-hand}(palm-involved) interactions, where a 
% stable palm frame is constructed from MCP joints to unexpected fingertip jitter; 
% (ii) \emph{Finger-only} interactions (pinch/precision), where the gripper axis aligns with the 
% index--thumb chord under a hand-intrinsic frame. In both cases we output a gripper rotation $R_g$ 
% and a position $\mathbf{p}_g$ with a small safety offset $d_z$ along the hand normal.

%%%%%%%%%%%%%%%%%%%%%%%
% \setlength{\columnseprule}{0pt}   % 去掉中间竖线

% \begin{figure*}[!h]
%     \centering
%     % \vspace{-1.5em}
%     \begin{overpic}[width=0.9\linewidth]{img/retargeting_gripper_v4.pdf}
%     \end{overpic}
%     \vspace{-1em}
%     \caption{}
%     \vspace{-1.8em}
%     \label{fig:pipeline}
% \end{figure*}

\vspace{-1em}

% \begin{figure*}[!h]
%     \centering
%     % \vspace{-1.5em}
%     \begin{overpic}[width=0.9\linewidth]{img/cross_domain_retargeting.pdf}
%     \end{overpic}
%     \vspace{-1em}
%     \caption{Retargeting to cross-domain embodiments}
%     \vspace{-1.8em}
%     \label{fig:pipeline}
% \end{figure*}

% \begin{figure} % "r" for right, adjust width as necessary
%     \centering
%     \vspace{-1.5em}
%     \begin{overpic}[width=\linewidth]{img/cross_domain_retargeting.pdf}
%     \end{overpic}
%     \vspace{-2em}
%     \captionsetup{font=footnotesize}
%     \caption{Cross-domain embodiment retargeting.}

%     \label{fig:cross_domain_retargeting}
% \end{figure}

\paragraph{Dexterous hands and humanoids.} 

Beyond retargeting to simple gripper-based arms, our high-fidelity HOI reconstruction offers a glimpse into transferring human demonstrations to more complex embodiments, including dexterous hands and humanoid robots.
For dexterous hands, we retarget the reconstructed hand motions to the joint space of target robotic hands using kinematic similarity and contact-preserving constraints. 
% This allows us to generate fine-grained finger motion trajectories that maintain functional grasp semantics.
For whole-body human demonstrations, we extend retargeting to humanoid platforms by leveraging full-body SMPL-H estimates. The resulting motion sequences are adapted to humanoid joint trees through inverse kinematics and dynamics-aware optimization, ensuring physical plausibility and intent preservation. Preliminary validations are provided in the Appendix.

\vspace{-1em}

\paragraph{Replay Assessment and Language Caption}
% Replay of trajectories on robotic arms in simulation inevitably leads to failures, especially when multiple data augmentation strategies are applied. At our dataset scale, fully manual triage is infeasible. Accordingly, we employ Qwen2.5-VL~\cite{bai2025qwen2} as an automatic evaluator, providing it with a brief task description and third-person observations from simulation as joint inputs to decide whether an episode succeeds. After filtering out successful episodes, we further refine their language instructions using ~\cite{hurst2024gpt} to support downstream robotic learning.

When replaying trajectories in simulation, executions may fail due to precision or non-physical collisions in the physics engine. To mitigate this, we employ Qwen2.5 VL~\cite{bai2025qwen2} for automatic binary task evaluation (success/failure). The model takes as input the task description along with both wrist- and third-view simulated observations to determine whether the action has been successfully completed. For trajectories deemed successful, we further derive and refine fine-grained language instructions using~\cite{hurst2024gpt}.

% Due to hardware constraints, we validated the retargeting results for both the dexterous hand and the humanoid robot exclusively in simulation, further revealing high-fidelity recovery of fine-grained contact topology and kinematic intent (see Appendix).

% With this unified retargeting framework, we expand the scalability and diversity of \nbname \ data engine. Each human demonstration is automatically transduced into multiple, semantically aligned training episodes spanning a broad range of robot embodiments—from parallel-jaw grippers to dexterous multi-fingered hands and humanoids. This mechanism materially amplifies the effective yield of every collected video by multiplying cross-embodiment supervision. By constructing a large-scale, cross-domain corpus in this manner, \nbname \ furnishes directly usable supervision for training generalist robotic policies that transfer skills and knowledge across heterogeneous hardware.

\vspace{-0.2em}

\subsection{Data Augmentation in Simulation}
\vspace{-0.3em}
%We enhance observation diversity in simulation through HOI-conditioned domain randomization while preserving the contact semantics essential for control. All HOI-to-workspace transformations are defined in the canonical action space \(\mathcal{A}\), thereby ensuring consistent contact frames and approach directions across randomized environments.

% While preserving the semantic and physical plausibility, we perform extensive simulation-based data augmentation to enhance the diversity of visual observations and motion trajectories, while maintaining the contact semantics critical for embodiment control. All augmentation strategies are applied within the canonical action space \(\mathcal{A}\).

While preserving contact semantics and physical plausibility that are critical for embodiment control, we perform extensive simulation-based data augmentation to enrich the diversity of visual observations and motion trajectories. All augmentation strategies are applied within the canonical action space \(\mathcal{A}\) to handle different manipulation orientations. More visualizations are provided in the Appendix.

% We enlarge observation diversity in simulation via HOI-conditioned domain randomization while preserving the contact semantics required for control. All transformers act in the canonical action space \(\mathcal{A}\), so that contact frames and approach directions remain consistent.

%An augmented rollout is admitted only if it passes a replay check consisting of end-effector tracking error \(d_{\mathrm{SE(3)}}(T_g(t),\hat T_g(t))<\epsilon_{\mathrm{ee}}\), zero penetration, and contact-IoU \(>\tau_{\mathrm{con}}\).
\vspace{-1.0em}

%%% Data Aug 1
\paragraph{Different types of arm retargeting.}

% Given an executable end-effector (EE) trajectory \(\{T_g(t)\}\) produced by our retargeting, we generate observations for heterogeneous arms. We instantiate five widely used 6–7\,DoF platforms: \emph{UR5/UR5e}, \emph{Franka Emika Panda}, \emph{KUKA LBR iiwa~7}, \emph{Kinova Gen3}, and \emph{UFactory xArm~7}. For each robot \(r\), we solve a rate-bounded IK with joint-limit and self-collision penalties:
% \begin{equation}
% \label{eq:ik}
% q_t^{(r)} \;=\; \arg\min_{q}\;
% d_{\mathrm{SE(3)}}\!\big(f^{(r)}_{\mathrm{FK}}(q),\,T_g(t)\big)
% \;+\;\lambda_{\mathrm{lim}}\phi_{\mathrm{lim}}(q)
% \;+\;\lambda_{\mathrm{sm}}\big\|\Delta^2 q\big\|_2^2,
% \end{equation}
% followed by time-scaling to satisfy velocity/acceleration bounds. 

\begin{figure*}[!h]
    \centering
    \vspace{-0.3em}
    \begin{overpic}[width=\linewidth]{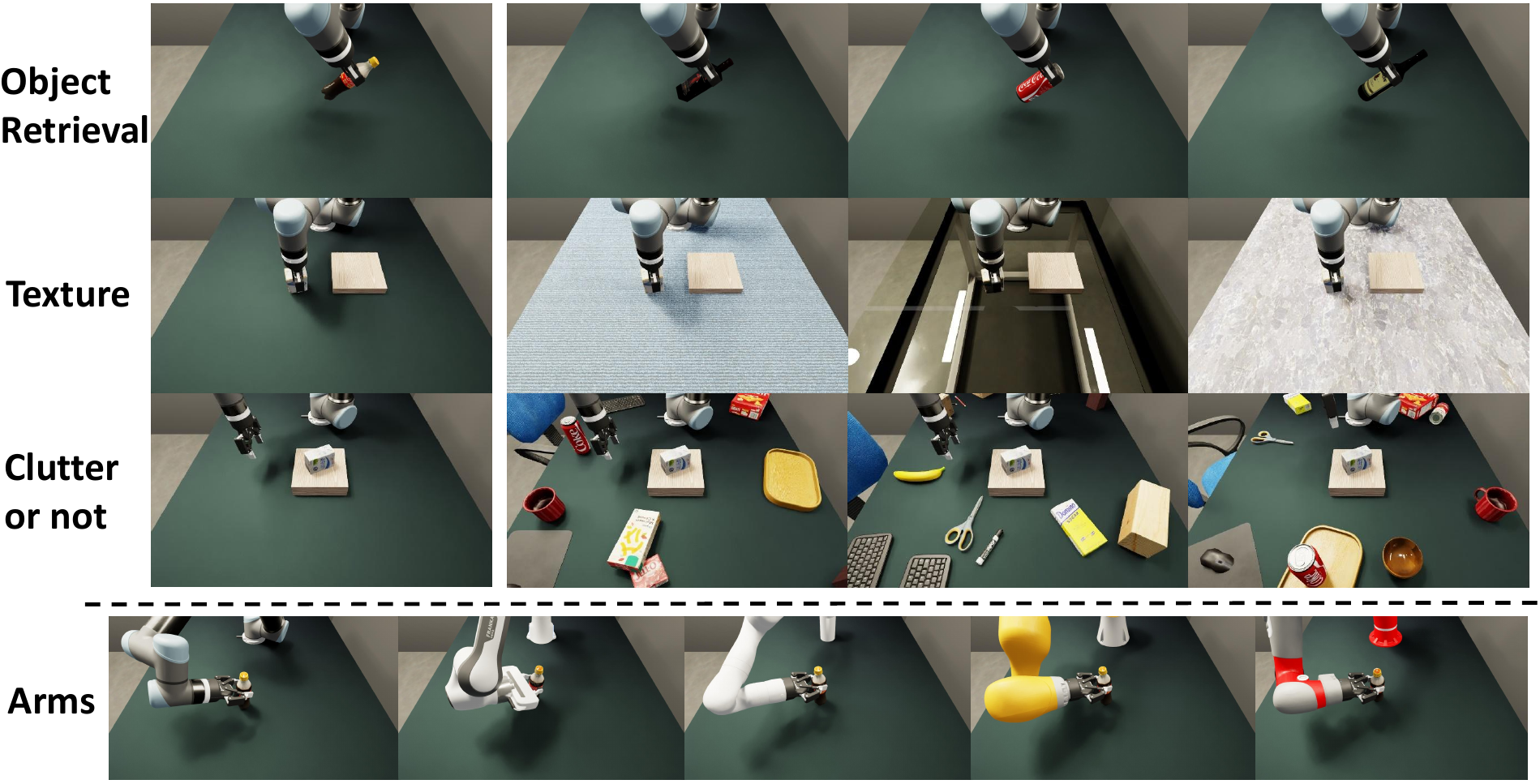}
    \end{overpic}
    \captionsetup{font=footnotesize}
    \caption{\nbname\,diverse augmentation in simulation.}
    \label{fig:handtoarms}
\end{figure*}

Given an executable end-effector (EE) trajectory \(\{T_g(t)\}\) produced by our retargeting method, we generate observations for heterogeneous arms, as illustrated in \cref{fig:handtoarms}. In Isaac Sim, we instantiate five widely used $6/7$-DoF robotic arms as simulation assets, including \emph{UR5/UR5e}, \emph{Franka Emika Panda}, \emph{KUKA LBR iiwa~7}, \emph{Kinova Gen3}, and \emph{Rethink Robotics Sawyer}. HOI-derived 6D EE trajectories \(T_g(t)\in\mathtt{SE}(3)\), \(t=1,\dots, T\), are mapped into feasible joint trajectories using cuRobo's GPU-accelerated inverse kinematic (IK) backend~\citep{sundaralingam2023curobo}.
For each robotic arm, at every timestep we invoke IK solver with the target pose \(T_g(t)\). The solver returns a feasible joint configuration:
\vspace{-1em}
% \begin{equation}
% \label{eq:ik_curobo}
% q_t \,=\, \arg\min_{q}  \mathcal{C}_{\text{goal}}(T_g(t), q) 
% \;\; 
% \text{s.t.} 
% q_{\min} \preceq q \preceq q_{\max}, 
% \mathcal{C}_{\text{coll}}(q) \le 0,
% \vspace{-0.5em}
% \end{equation}

\begin{equation}
\label{eq:ik_curobo}
\begin{aligned}
q_t \,=\, \arg\min_{q}  \mathcal{C}_{\text{goal}}(T_g(t), q) \\
\text{s.t.} 
q_{\min} \preceq q \preceq q_{\max},\;
\mathcal{C}_{\text{coll}}(q) \le 0,
\end{aligned}
\vspace{-0.3em}
\end{equation}
where $\mathcal{C}_{\text{goal}}$ is cuRobo’s pose reaching cost and $\mathcal{C}_{\text{coll}}$ is the self-collision constraint. To encourage temporal consistency, we use the previous solution $q_{t-1}$ as the IK seed when invoking the solver.

% \begin{equation}
% \begin{aligned}
% \min_{q} \quad & \mathcal{C}_{\text{goal}}(T_g(t), q) + \lambda \Vert q - q_{t-1} \Vert_2^2 \
% \text{s.t.} \quad & q_{\min} \preceq q \preceq q_{\max} \
% & \mathcal{C}{\text{self}}(q) \le 0
% \end{aligned}
% \end{equation}

% \begin{equation}
% \label{eq:cand_gen_updated}
% q_{t} \;=\; \arg\min_{q\in\mathcal{Q}_t}\;
% d_{\mathrm{SE(3)}}\!\big(f_{\mathrm{FK}}(q),\,T_g(t)\big)
% +\lambda_{\mathrm{lim}}\phi_{\mathrm{lim}}(q)
% +\lambda_{\mathrm{col}}\phi_{\mathrm{col}}(q),
% \end{equation}
% where \(f_{\mathrm{FK}}\) is the forward kinematics, \(d_{\mathrm{SE(3)}}\) measures pose error, and \(\phi_{\mathrm{lim}},\phi_{\mathrm{col}}\) denote joint-limit and self-collision penalties. 
% Here \(\mathcal{Q}_t\) denotes the candidate joint-space configurations considered by the IK solver at timestep \(t\), constrained by the robot’s kinematic limits and search heuristics.

% To reconcile differences between the HOI end-effector frame and the robot's nominal tool frame, we apply a fixed, robot–gripper specific TCP transform \(\Delta T_{\mathrm{tcp}}^{(r)}\):
% \begin{equation}
% \label{eq:target_pose}
% T_g^{(r)}(t) \;=\; T_g(t)\circ\Delta T_{\mathrm{tcp}}^{(r)},
% \end{equation}
% where \(\circ\) denotes composition in \(\mathrm{SE(3)}\). The transform \(\Delta T_{\mathrm{tcp}}^{(r)}\) accounts for both mechanical offsets of the attached gripper and task-specific adjustments. For instance, when the EE pose is palm-induced, we keep the approach axis and opening width and add a safety lift \(d_z\) along the palm normal to mitigate fingertip jitter.
Episodes that pass the replay check preserve the original HOI intent (\textit{e.g.}, grasp/place/pour) while providing embodiment diversity in joint space. We export both the joint-space commands \(\{q_t\}_{t=1}^T\) (arm and gripper included) and aligned operational-space labels per robot, enabling multi-morphology policy training from the same HOI source.

%%% Data Aug 2

\paragraph{Object retrieval and replacement.} 
We build a large object library by combining~\cite{hunyuan3d2} generations with in-house scans; each asset includes a watertight mesh, texture, category tag, and a canonical pose. For a source episode with object mesh \(M_o\) and object pose stream \(\{T_o(t)\}\), we retrieve top-\(K\) substitutes \(\tilde M\) = \(\{\tilde M_k\}\) using a fused similarity, 
    \vspace{-0.5em}
\begin{equation}
    \small
    \begin{aligned}        
    \mathcal{S}(M_o,\tilde M)
    =&\alpha\,\mathrm{CD}\!\big(\hat M_o,\hat{\tilde M}\big)
    +\beta\,\big(1-\mathrm{IoU}_{\mathrm{AABB}}\big)\\
    &+\gamma\,\big\langle \phi_{\mathrm{sem}}(M_o),\,\phi_{\mathrm{sem}}(\tilde M)\big\rangle,
    \end{aligned}
        \vspace{-0.8em}
\label{eq:retrieval}
\end{equation}
where \(\hat{\cdot}\) denotes unit AABB normalization, \(\mathrm{CD}\) is the symmetric Chamfer distance on surface samples, \(\mathrm{IoU}_{\mathrm{AABB}}\) measures coarse shape compatibility, and \(\phi_{\mathrm{sem}}\) are text–shape embeddings.

To ensure replay compatibility on a retrieved substitute, we align principal axes and bind the same maximum AABB and canonical pose definition as the source \(M_o\). Under this binding, the original EE motion plan and the hand–object interaction geometry remain consistent, so the control trajectory can be directly replayed on geometrically/semantically matched novel objects (\textit{e.g.}, mug \(\!\leftrightarrow\!\) cup-with-handle, box \(\!\leftrightarrow\!\) carton).

\vspace{-1.5em}

\paragraph{Trajectory augmentation.} 
Informed by \cite{xue2025demogen} and tailored to our setting, we represent each demonstration as a trajectory $\tau=\{(T_g(t),g(t))\}_{t=1}^T$, where $T_g(t)=(R(t),p(t))\in\mathtt{SE}(3)$ denotes the EE pose with orientation $R(t)$ and translation $p(t)$, and $g(t)$ is the gripper command. The trajectory is partitioned into object-centric segments $\{\tau^{(k)}\}$, each labeled by a contact state $c^{(k)}\in\{\texttt{hold},\texttt{open}\}$. Instead of re-planning trajectories, we augment them as follows. 

\textit{(i)} For interaction segments ($c^{(k)}=\texttt{hold}$), we apply an object-frame rigid transform $T_o\in\mathtt{SE}(3)$ to each waypoint:
\vspace{-0.6em}
\[
\tilde T_g(t) = T_o T_g(t),\qquad \tilde g(t) = g(t).
    \vspace{-0.3em}
\]
Let $R_\Delta := \mathtt{Rot}(T_o)$. To maintain continuity without motion-plan regeneration, the same EE orientation change is applied to non-interaction segments (see (ii)), and the orientation change induced by $R_\Delta$ is kept small for IK feasibility and repeatable execution.

\textit{(ii)} For each non-interaction segment ($c^{(k)}=\texttt{open}$), we linearly remap the translational path and set the EE orientation as $\tilde R(t) = R_\Delta R(t)$. Let $p_s,p_e$ be the original endpoints and $\hat p_s,\hat p_e$ the remapped anchors: the anchor adjacent to a transformed interaction segment is fixed by that segment, while the opposite anchor is chosen within a predefined reachable set. With $\alpha_t\in[0,1]$ denoting the normalized progress along the original segment from $p_s$ to $p_e$,

\vspace{-1.7em}

% \[
% \tilde p_t = \hat p_s + \alpha_t(\hat p_e - \hat p_s) +\Big[p_t - \big(p_s + \alpha_t(p_e - p_s)\big)\Big].
% \]
\begin{equation}
\tilde p_t \;=\; \hat p_s \;+\; \alpha_t(\hat p_e - \hat p_s)
\;+\;\Big[p_t - \big(p_s + \alpha_t(p_e - p_s)\big)\Big].
\end{equation}

\section{Dataset}
\label{sec:dataset}
\vspace{-0.3em}

Using the \nbname \ pipeline, we build a large-scale multimodal HOI-robotic dataset \textbf{\nbdata}. The dataset integrates multiple data sources and leverages successive stages of the \nbname \ to convert heterogeneous inputs into a unified hoi representation, followed by cross-embodiment retargeting and diversified data augmentation. Data sources include: (i) a custom multi-view mocap system equipped with tactile sensor–instrumented gloves, (ii) multiple public HOI datasets, and (iii) a custom RGB(D) video capture setup. For source (i), high-precision HOI signals are obtained via triangulation and FoundationStereo~\cite{wen2025foundationstereo}, while the remaining robot-relevant modalities are generated using the \nbname \  retargeting and augmentation pipeline. For source (ii), existing HOI modalities in public datasets are directly convert to the canonical action space $\mathcal{A}$, then retargeted and augmented in the same manner. For source (iii), the full \nbname \ pipeline processes raw videos to extract both HOI and embodied modalities. 
For detailed information on the tasks included in each subset, please refer to our Appendix.
% The following sections detail the processing steps and included modalities for each subset.
The following sections detail the modality composition and statistics of \nbdata \ , as well as the acquisition system for the \textbf{mocap subset}.

\vspace{-0.5em}
% The resulting corpus, which we term \textbf{RoboWheel-150K}, contains approximately 150\,000 trajectories spanning diverse objects, tasks, viewpoints, and robot embodiments. The data is generated by running the three-stage pipeline in Sec.~\ref{sec:method} on three types of sources:

\subsection{Modalities Composition and Statistics.}
%Across all sources, each episode is represented in a unified, metrically consistent format. Concretely, we provide synchronized visual observations, hand and object pose, contact state, and robot control labels.
%As summarized in Tab.~\ref{tab:rw1m_comparison}, \nbname\ produces a unified multimodal representation across three subsets of \nbdata 
%: a mocap subset, an RGB(D) recording subset, and a subset derived from public HOI datasets. Each episode is cast into a common HOI–robotic format, but the available modalities differ slightly across subsets. 
As summarized in Tab.~\ref{tab:rw1m_comparison}, \nbname\ produces a unified multimodal representation across the three subsets of \nbdata. These modalities can be broadly grouped into two categories: those for HOI downstream tasks and those for robot-related tasks. The HOI modalities include sequences of hand MANO parameters (pose/shape with global orientation and translation) in world frame from the original data source, object 6dof pose, object assets, and hand–object contact annotations. The robot-related modalities include observations from the robot wrist and third-person viewpoints, as well as end-effector pose trajectories for the robotic arm. All trajectories are transformed into the canonical action space $\mathcal{A}$.

The \textbf{mocap subset} provides tactile signals, robotic observations, and HOI annotations. Concretely, we release the 6-DoF object pose, object assets, and a dense tactile map for both the hand and the object. The \textbf{recording subset} provides the same HOI and robotic modalities, except for tactile signals, because the collection setup did not include the hardware. The \textbf{public HOI subset} is constructed from existing HOI datasets~\cite{taheri2020grab,hampali2021ho3dv3,Chao_2021_CVPR,wang2024ho,li2023taco} and currently includes the cross-embodiments modalities retargeted to different arms ($6/7$-DoF). The three subsets together constitute roughly 150k trajectories, distributed shown in Fig.~\ref{fig:robowheel_pie}.

\begin{table*}[h]
\centering
\begin{minipage}{0.73\textwidth}
    % \caption{Modalities and scale comparison. For \nbdata, the first three rows correspond to our mocap, recorded RGB(D), and public HOI subsets. ``\ding{55}'' and ``\cmark'' denote absence and presence, respectively.}
    %\label{tab:rw1m_comparison}

    \renewcommand{\arraystretch}{1.2}
    \setlength{\tabcolsep}{2pt}

    % colors sampled from the pie chart
    % \definecolor{rwMocap}{RGB}{112,112,212}     % light bluish purple
    % \definecolor{rwRecording}{RGB}{ 64, 65,122} % dark blue
    % \definecolor{rwPublic}{RGB}{245,241,230}    % cream

    \definecolor{rwMocap}{RGB}{64, 65,122}     % light bluish purple
    \definecolor{rwRecording}{RGB}{245,241,230} % dark blue
    \definecolor{rwPublic}{RGB}{112,112,212}    % cream

    % editable background choices (tinted for readability)
    \newcommand{\rwMocapBg}{rwMocap!35}
    \newcommand{\rwRgbBg}{rwRecording!35}
    \newcommand{\rwPublicBg}{rwPublic!80}

    \resizebox{0.95\linewidth}{!}{%
    \begin{tabular}{l
                    >{\centering\arraybackslash}p{1.6cm}
                    >{\centering\arraybackslash}p{1.5cm}
                    >{\centering\arraybackslash}p{2cm}
                    >{\centering\arraybackslash}p{2.2cm}
                    p{6.3cm}}
    
    \toprule
    \thead{Dataset} &
    \thead{Tactile} &
    \thead{Robotic data} &
    \thead{HOI data} &
    \thead{\# Trajectories} &
    \thead{Object Info Granularity} \\
    \midrule
    GRAB~\cite{taheri2020grab} & \ding{55} & \ding{55} & \cmark & 1334 &
    6-DoF pose \& Contact maps \\
    HO3D (v3)~\cite{hampali2021ho3dv3} & \ding{55} & \ding{55} & \cmark & 68 &
    6-DoF pose \& Assets \\
    DexYCB~\cite{Chao_2021_CVPR} & \ding{55} & \ding{55} & \cmark & 1{,}000 &
    6-DoF pose \& Assets \\
    HO-Cap~\cite{wang2024ho} & \ding{55} & \ding{55} & \cmark & $\sim$64 &
    6-DoF pose \& Assets \\
    TACO~\cite{li2023taco} & \ding{55} & \ding{55} & \cmark & 2{,}500 &
    Assets \\
    \midrule
    DROID~\cite{khazatsky2024droid} & \ding{55} & \cmark & \cmark & 76{,}000 &
    Assets \\
    LIBERO~\cite{liu2023libero} & \ding{55} & \cmark & \ding{55} & 366 &
    Assets \\
    UCSD Kitchen~\cite{ucsd_kitchens} & \ding{55} & \cmark & \ding{55} & 150 &
    $ \xmark $ \\
    \midrule
    \rowcolor{\rwRgbBg}
    \textbf{\nbdata (Recordings)} &
    \ding{55} & \cmark & \cmark & 23560 &
    6-DoF object pose \& Assets \\
    \rowcolor{\rwMocapBg}
    \textbf{\nbdata (Mocap)} &
    \cmark & \cmark & \cmark & 63141 &
    6-DoF object pose \& Assets \& Tactile map \\
    \rowcolor{\rwPublicBg}
    \textbf{\nbdata (Public Dataset)} &
    \ding{55} & \cmark & \cmark & 66924 &
    \xmark \\
    \bottomrule
    \end{tabular}}
    \caption{Modalities and scale comparison. For \nbdata, the first three rows correspond to our mocap, recorded RGB(D), and public HOI subsets. ``\ding{55}'' and ``\cmark'' denote absence and presence, respectively.}
    \label{tab:rw1m_comparison}
\end{minipage}
\hfill
% ---------- Figure (pie chart) ----------
\begin{minipage}{0.24\textwidth}
    \captionsetup{type=figure}
    \centering
    \includegraphics[width=\linewidth]{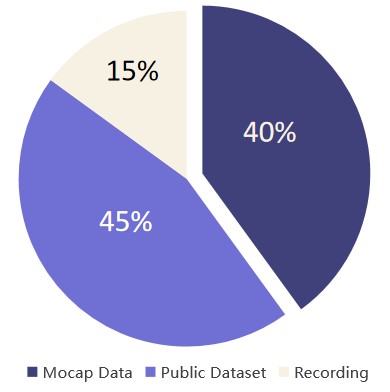}
    \vspace{-1em}
    \caption{Composition of the three subsets in \nbdata.}
    \label{fig:robowheel_pie}
\end{minipage}
\vspace{-1em}
\end{table*}

% Each episode includes: (a) synchronized RGB observation, (b) per-frame MANO parameters in world frame, (c) 6-DoF object pose together with a textured mesh, (iv) interaction state between the robot and objects, (v) language description of the task and object-level goal.

\begin{figure}[!h]
    \centering
    \vspace{-0.7em}
    \begin{overpic}[width=0.9\linewidth]{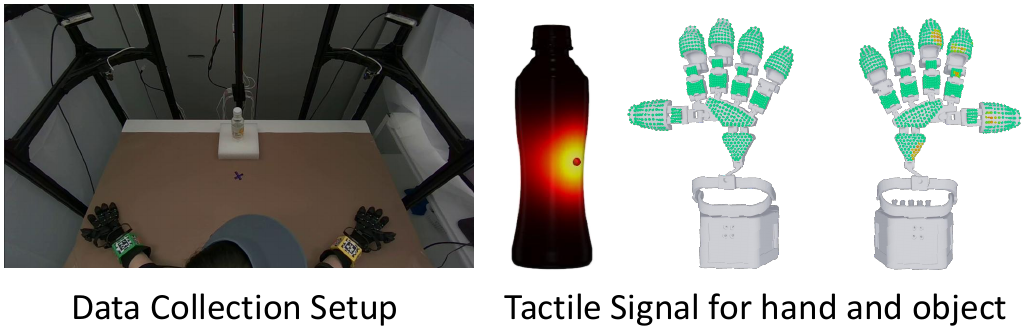}
    \end{overpic}
    \vspace{-1em}
    \caption{Data collection setup and tactile information.}
    \label{fig:tactile}
    \vspace{-1.5em}
\end{figure}

%\nbdata \, dataset contains approximately 150k trajectories drawn from multiple data sources, reprocessed public HOI corpora, and our studio captures. Each episode includes synchronized multi-view RGB/RGB-D observations, per-frame MANO parameters in the world frame (pose/shape with global orientation and translation), 6-DoF object pose with a textured asset (mesh/texture ID), contact states and partially available tactile signals(shown in ~\cref{fig:tactile}), and fine-grained language task descriptions. For more details, please refer to the Appendix.

\subsection{Custom-built Mocap system}
\paragraph{Acquisition System}
\vspace{-0.3em}

Our high-precision mocap subset is collected with a synchronized, multi-sensor capture rig, the data acquisition setup and tactile glove are shown in ~\cref{fig:tactile}. We use three Intel RealSense D455 RGB-D cameras connected via hardware sync cables to obtain time-aligned, metric depth from multiple viewpoints, and eight additional RGB cameras, yielding up to eleven views for each interaction. All cameras are extrinsically calibrated into a common world coordinate frame and temporally synchronized with the hand instrumentation.

Hand motion and tactile sensing are recorded using \textbf{EVT2} developed by \textbf{Paxini}. The glove integrates
\begin{itemize}
    \item \textbf{29 magnetic encoders} distributed over the hand kinematic chain (wrist, palm, and phalanges), which provide joint-level pose measurements.
    \item \textbf{16 Gen3 tactile (haptic) sensors} mounted on key contact regions (fingertips, finger pads, and palm), which measure local normal forces.
\end{itemize}
These signals are used both for fitting MANO parameters and for deriving tactile and contact annotations that we release as part of the dataset.

\vspace{-1.3em}

\paragraph{Fitting to MANO with tactile signals}
% \paragraph{Glove-to-MANO/URDF Mapping and Contact/Tactile Signals}
% We develop a tactile-guided, multi-constraint optimization scheme for motion retargeting. By registering glove measurements to the MANO hand model (or a target robot URDF) and jointly enforcing constraints on tactile contact states, wrist floating calibration, joint smoothness, and hand anatomical priors, the system optimizes all frames in parallel to produce spatiotemporally consistent, tactile-synchronized, and physically plausible motion trajectories.
We develop a tactile-guided, multi-constraint optimization scheme for motion retargeting. By registering glove measurements to the MANO hand model (or a target robot URDF) and jointly enforcing constraints on tactile contact states, wrist pose calibration, joint smoothness, and anatomical priors, the system optimizes all frames in parallel to produce spatiotemporally consistent, tactile-synchronized, and physically plausible motion trajectories. See Appendix for implementation details of the retargeting algorithm.

    \vspace{-0.5em}
\section{Experiment}
    \vspace{-0.3em}

\subsection{HOI Reconstruction Quality}
\begin{figure*}[!h]
    \centering
    \begin{overpic}[width=0.95\linewidth]{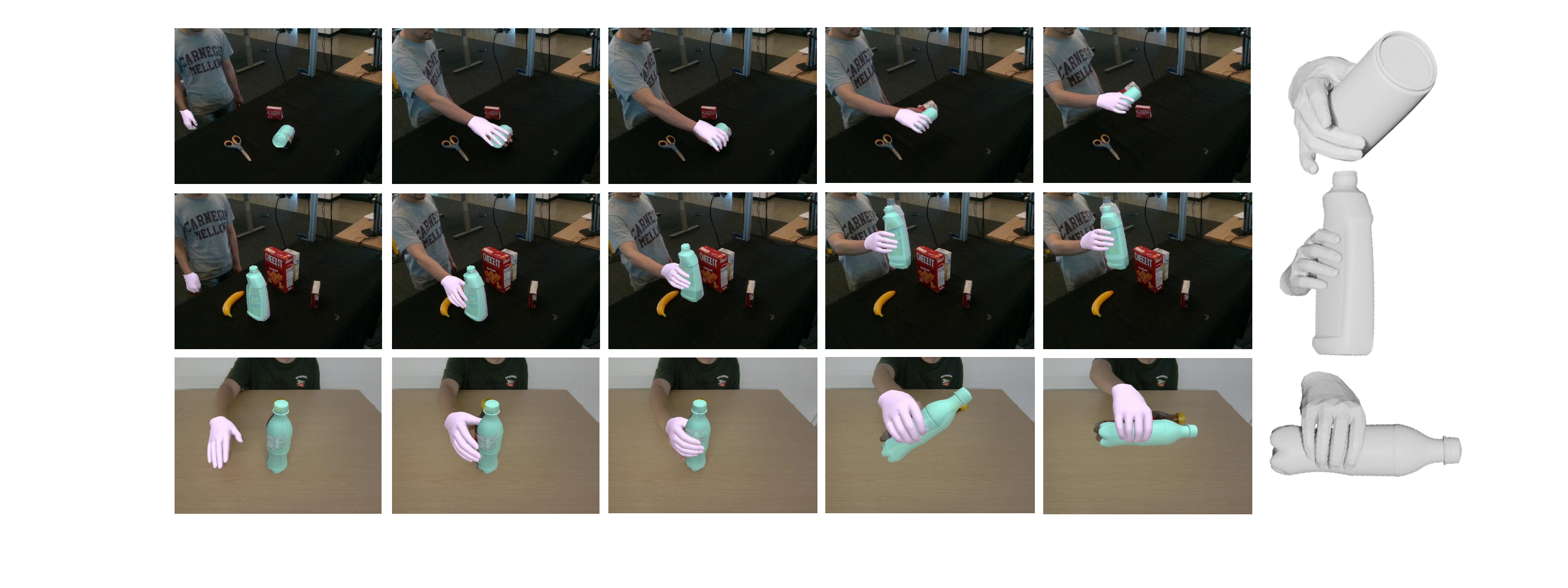}
    \end{overpic}
    \caption{HOI reconstruction results of \nbname \ .Whether the data comes from public HOI datasets (\textit{e.g.}, DexYCB) or not, \nbname \ can achieve high-precision HOI reconstruction.}
    \label{fig:hoi}
\end{figure*}

    \vspace{-0.3em}
We evaluate HOI reconstruction quality on~\cite{wang2024ho} with common metrics in ~\cref{tab:hoi-recon-quality1} and ~\cref{fig:hoi}. All methods receive the same camera parameters and object meshes for a fair comparison. We compare \nbname \ with HORT~\citeyearpar{chen2025hort}, HOLD~\citeyearpar{zhao2023aloha}, and DiffHOI~\citeyearpar{ye2023diffusion}.

\begin{table*}[!h]
  \centering
  % \small
  \setlength{\tabcolsep}{4.6pt}

    \definecolor{HOIours}{RGB}{200, 200,200}     % light bluish purple
    \newcommand{\HOIours}{HOIours}

    %\captionsetup{font=footnotesize}
    % \caption{HOI reconstruction quality comparison.
    % \textbf{Object surface:} CD (cm) = bidirectional Chamfer distance; F5/F10 (\%) = F-score at 5/10\,mm.
    % \textbf{Hand:} Hand jitter (cm/s$^2$) = time-avg.\ norm of frame-to-frame wrist/palm acceleration (30\,FPS, 2nd-order diff.);
    % W-MPJPE (mm) = wrist-relative MPJPE after aligning the wrist.
    % \textbf{Rel. pose consistency:} std of $T_{\mathrm{rel}}(t)=T_h^{-1}(t)T_o(t)$ in translation (cm) / rotation (deg).}
    % \label{tab:hoi-recon-quality1}
        % \label{tab:hoi-recon-quality}
  \resizebox{\linewidth}{!}{
  \begin{threeparttable}
    \begin{tabular}{
      % l
      % S[table-format=1.2]  % CD (cm)
      % S[table-format=2.1]  % F5 (%)
      % S[table-format=2.1]  % F10 (%)
      % S[table-format=1.2]  % Hand jitter (cm/s^2)
      % S[table-format=2.1]  % WPJPE (mm)
      % S[table-format=1.2]  % Rel trans std (cm)
      % S[table-format=1.1]  % Rel rot std (deg)
          l
          c c c
          c c
          c c
    }
    \toprule
    \textbf{Method}
      & \multicolumn{3}{c}{\textbf{Object}} 
      & \multicolumn{2}{c}{\textbf{Hand}} 
      & \multicolumn{2}{c}{\textbf{Rel. pose consistency} $\downarrow$} \\
    \cmidrule(lr){2-4}\cmidrule(lr){5-6}\cmidrule(lr){7-8}
    {}
      & \multicolumn{1}{c}{CD (cm)$\downarrow$}
      & \multicolumn{1}{c}{F5 (\%)$\uparrow$}
      & \multicolumn{1}{c}{F10 (\%)$\uparrow$}
      & \multicolumn{1}{c}{Hand jitter (cm/s$^2$)$\downarrow$}
      & \multicolumn{1}{c}{W-MPJPE (mm)$\downarrow$}
      & \multicolumn{1}{c}{Trans (cm)$\downarrow$}
      & \multicolumn{1}{c}{Rot (deg)$\downarrow$} \\
    \midrule
      HORT          & 8.9 & 55.0 & 83.0 & 3.35 & 19.92 & 3.54 & - \\
      DiffHOI       & 7.2 & 59.6 & 78.1 & 4.59 & 20.21 & 4.51 & - \\
      HOLD          & 7.5 & 53.2 & 77.9 & 3.47 & 20.59 & 2.44 & - \\
    \rowcolor{HOIours}
    \textbf{Ours}   & \textbf{5.1} & \textbf{63.4} & \textbf{89.1} & \textbf{0.92} & \textbf{7.81} & \textbf{0.26} & \textbf{1.9} \\
      \bottomrule
    \end{tabular}
    
    \label{tab:hoi-recon-quality1}
  \end{threeparttable} 
  }
  \vspace{-1em}
    \caption{HOI reconstruction quality comparison.
    \textbf{Object surface:} CD (cm) = bidirectional Chamfer distance; F5/F10 (\%) = F-score at 5/10\,mm.
    \textbf{Hand:} Hand jitter (cm/s$^2$) = time-avg.\ norm of frame-to-frame wrist/palm acceleration (30\,FPS, 2nd-order diff.);
    W-MPJPE (mm) = world frame MPJPE.
    \textbf{Rel. pose consistency:} std of $T_{\mathrm{rel}}(t)=T_h^{-1}(t)T_o(t)$ in translation (cm) / rotation (deg).}
\vspace{-0.3em}
\end{table*}

\vspace{-0.3em}
\subsection{ Validation of \textbf{\nbdata} Data }
\vspace{-0.5em}

Here we show how \nbdata \ data support downstream tasks and how well learned skills transfer with data augmentation. 
\vspace{-2.5em}

\paragraph{Performance on different VLA/IL models.}
To study how Robowheel reconstructions translate to downstream control, we benchmark eight household tasks grouped by difficulty (Easy/Hard) and evaluate several VLA/IL algorithms (ACT, DP, RDT, Pi0). We evaluate each algorithm under three training regimes: (i) fine-tuning on 10 teleoperation demonstrations (tele.), (ii) fine-tuning on 10 \nbdata \,trajectories (\nbdata), and (iii) a two-stage curriculum that first pre-trains on 5k \nbdata \,trajectories and then fine-tunes on 10 \nbdata \,trajectories (reported as RDT+5k\nbdata \  and Pi0+5k\nbdata ).The third training regime is applied only to RDT and Pi0, as the other methods do not support pre-training. Per-task success rate(\%)  is reported in the same real setup in~\cref{tab:vla-ml}.
Macro averages within each difficulty group are reported. See Appendix for more results.

%\newcolumntype{L}[1]{>{\raggedright\arraybackslash}p{#1}}

\begin{table*}[!t]

\centering
\setlength{\tabcolsep}{2pt}      

\definecolor{VLAours}{RGB}{200, 200,200}     % light bluish purple
\newcommand{\VLAours}{VLAours}

\renewcommand{\arraystretch}{1} 
\scriptsize                      
\resizebox{\linewidth}{!}{
\begin{tabular}{
% L{2.1cm} c *{6}{c}
 L{2.1cm}  % 任务名
  c         % 难度
  c c c c   % 前四个方法
  >{\columncolor{\VLAours}}c   % 倒数第二列：RDT+5kHO
  >{\columncolor{\VLAours}}c 
} % 8 列：左列自适应换行 + 1 列难度 + 6 列结果
\toprule
\multirow{2}{*}{\textbf{Real-world Tasks}} & \multirow{2}{*}{\textbf{Diff.}} 
& \makecell{ACT\\tele.\,\textbar{} \,\nbdata} 
& \makecell{DP\\tele.\,\textbar{} \,\nbdata}
& \makecell{RDT\\tele.\,\textbar{} \,\nbdata}
& \makecell{Pi0\\tele.\,\textbar{} \,\nbdata}
& \makecell{RDT+5k\nbdata\\\nbdata}
& \makecell{Pi0+5k\nbdata\\\nbdata} \\
\midrule
Pick up milk& \multirow{4}{*}{Easy}  
& 15.0 \textbar{} 0.0 \;& 20.0 \textbar{} 15.0 & 55.0 \textbar{} 30.0 & 70.0 \textbar{} 65.0 & \textbf{70.0}  &  \textbf{80.0} \\

Lift wooden cup &  
& 0.0 \textbar{} 0.0 & 45.0 \textbar{} 25.0 & 75.0 \textbar{} 45.0 & 65.0 \textbar{} 55.0 &  \textbf{70.0} &  \textbf{70.0} \\

Place milk & 
& 35.0 \textbar{} 0.0 \;& 50.0 \textbar{} 35.0 & 70.0 \textbar{} 50.0 & 80.0 \textbar{} 55.0 &  \textbf{85.0} & \textbf{80.0} \\

Restore bowl & 
& 0.0 \textbar{} 0.0 & 5.0 \textbar{} 0.0 & 65.0 \textbar{} 65.0 & 60.0 \textbar{} 60.0 &  \textbf{75.0}  &  \textbf{75.0} \\
\midrule
Average & & 12.5 \textbar{} 0.0 \;& 30.0 \textbar{} 18.8 & 66.3 \textbar{} 47.5 & 68.8 \textbar{} 58.8 & \textbf{75.0} &  \textbf{76.3} \\
\midrule

Move banana& \multirow{4}{*}{Hard}     
& 0.0 \textbar{} 0.0 & 5.0 \textbar{} 0.0 & 55.0 \textbar{} 20.0   & 40.0 \textbar{} 15.0  &  \textbf{65.0}  & \textbf{60.0} \\

Upright milk  & 
& 0.0 \textbar{} 0.0 &  \; 0.0 \textbar{} 15.0 & 45.0 \textbar{} 30.0 & 60.0 \textbar{} 50.0 & \textbf{60.0} &  \textbf{75.0} \\

Pour cola & 
& 0.0 \textbar{} 0.0 & \; 0.0 \textbar{} 10.0 & 35.0 \textbar{} 35.0 & 25.0 \textbar{} 35.0 & \textbf{40.0}  &  \textbf{55.0} \\

Tip teacup  & 
& 0.0 \textbar{} 0.0 & 0.0 \textbar{} 0.0 & \; 5.0 \textbar{} 15.0 & 35.0 \textbar{} 25.0 & \textbf{25.0} &  \textbf{45.0} \\
\midrule
Average & & 0.0 \textbar{} 0.0 & 1.3 \textbar{} 6.3 & 35.0 \textbar{} 25.0 & 40.0 \textbar{} 31.3 & \textbf{47.5} &  \textbf{58.8} \\
\bottomrule
\end{tabular}

}% end scalebox

\caption{Real-world task performance (\%) grouped by difficulty.}
\label{tab:vla-ml}
\end{table*}
% 如需恢复默认：
% \setlength{\tabcolsep}{6pt}\renewcommand{\arraystretch}{1.0}

\begin{figure*}[!h]
    \centering
    \begin{minipage}[h]{0.68\linewidth}
        \centering
        \setlength{\tabcolsep}{8pt}
        \renewcommand{\arraystretch}{1.2}
        %\captionof{table}{Task performance in unseen scenarios with RDT.}
        % \label{tab:ablation_RDT_unseen}
        \resizebox{\textwidth}{!}{
                \tiny
        \begin{tabularx}{0.95\textwidth}{@{}l *{6}{>{\centering\arraybackslash}X}@{}}
        \toprule
        \multirow{2}{*}{\textbf{Real-world Tasks}} 
        & \multicolumn{2}{c}{\textbf{Unseen Object}} 
        & \multicolumn{2}{c}{\textbf{Clutter Objects}} 
        & \multicolumn{2}{c}{\textbf{Unseen Background}} \\
        \cmidrule(lr){2-3} \cmidrule(lr){4-5} \cmidrule(lr){6-7}
        & \textbf{\nbdata} & \textbf{\nbdata-aug} 
        & \textbf{\nbdata} & \textbf{\nbdata-aug} 
        & \textbf{\nbdata} & \textbf{\nbdata-aug} \\
        \midrule
        Lift wooden cup & 4/10 & \textbf{5/10} & 4/10 & \textbf{4/10} & 0/10 & \textbf{4/10} \\
        Place milk      & 5/10 & \textbf{6/10} & 5/10 & \textbf{6/10} & 1/10 & \textbf{3/10} \\
        Upright milk    & 3/10 & \textbf{3/10} & 4/10 & \textbf{4/10} & 3/10 & \textbf{5/10} \\
        Pour cola       & 3/10 & \textbf{3/10} & 3/10 & \textbf{4/10} & 2/10 & \textbf{4/10} \\
        \midrule
        \textbf{Average} & 3.75/10 & \textbf{4.25/10(+0.5)} & 4.00/10 & \textbf{4.50/10(+0.5)} & 1.50/10 &\textbf{ 4.00/10(+2.5)} \\
        \bottomrule
        \end{tabularx}
        }

        \captionof{table}{Task performance in unseen scenarios with RDT.}
        \label{tab:ablation_RDT_unseen}
    \end{minipage}
    \hfill
    % ---- 左：超紧凑表格 ----
    \begin{minipage}[h]{0.28\linewidth}
        % \captionof{table}{Robot \textbf{direct replay} success rate $\uparrow$ (\%).}
        % \label{tab:realrobotreplay}
        \vspace{-1em}
        \centering
        \scriptsize
        \setlength{\tabcolsep}{3pt}            % 缩小列间距
        \definecolor{replayours}{RGB}{200, 200,200}     % light bluish purple
        \newcommand{\replayours}{replayours}
        \renewcommand{\arraystretch}{0.95}     % 略微压缩行高
                \resizebox{0.9\textwidth}{!}{
        \begin{threeparttable}           
            \begin{tabular}{
            % @{}l *{3}{S[table-format=3.1]}@{}
            @{}
            l
            >{\columncolor{\replayours}}S[table-format=3.1]
            *{2}{S[table-format=3.1]}
            @{}
            } % 去除左右边距 @{}
              
              \toprule
              \textbf{Task} & {\textbf{Ours}} & {\textbf{GAT-Grasp}}  & {\textbf{yoto}} \\
              \midrule
              Lift wooden cup & \textbf{87.5}  & {25.0} & {75.0} \\
              Place bowl      & \textbf{100}   & {62.5} & {100} \\
              Upritght milk   & \textbf{100} & {50.0} & {62.5} \\
              Pour cola       & \textbf{100}   & {62.5} & {75.0} \\
              move banana     & \textbf{87.5}  & {62.5} & {87.5} \\
              Tip teacup      & \textbf{87.5}  & {37.5} & {00.0} \\
              \midrule
              \textbf{Macro avg} & \textbf{91.7} & {50.0} & {66.7} \\
              \bottomrule
            \end{tabular}
            \captionof{table}{Robot \textbf{direct replay} success rate $\uparrow$ (\%).}
            \label{tab:realrobotreplay}
        \end{threeparttable}
        }
    \end{minipage}

\end{figure*}

\begin{figure*}[!h]
    \centering
    \hspace{-0.6em}
    \begin{overpic}[width=\linewidth]{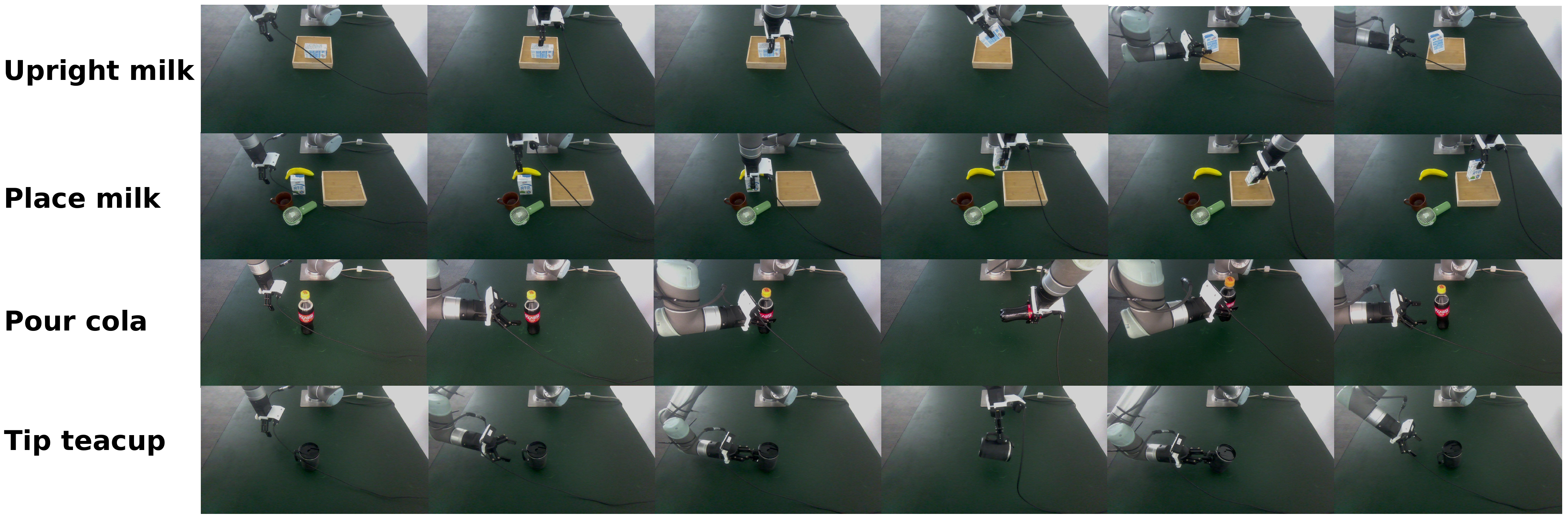}
    \end{overpic}
    \vspace{-2.2em}
    \caption{Real-world Performance on 4 Tasks.}
    \vspace{-1.7em}
    \label{fig:real robot main}
\end{figure*}
As shown in ~\cref{tab:vla-ml,fig:real robot main}, VLA policies (Pi0 \& RDT) pretrained with 5k \nbdata{}\ data achieved the highest success rates, demonstrating a remarkable performance improvement over counterparts without pretraining. The impact of this improvement is more evident in tasks of higher complexity. Notably, when training or fine-tuning these VLA/IL methods with an identical number of training episodes from teleoperation or \nbdata, policies trained solely on \nbdata \  achieve performance comparable to those trained on teleoperation data, despite the sim-to-real gap. 
% This result highlights the effectiveness of \nbname{} data for augmentation.
The underlying reason for this is twofold. First, \nbname\, provides precise HOI reconstruction, enabling the generation of trajectories with accuracy approaching that of teleoperation, ensuring effective transfer from simulation to real-world environments. Second, data augmentation is applied to \nbdata{}, and its broader data distribution helps the policy mitigate the negative impact of the visual domain gap, endowing the policies with enhanced robustness.

\vspace{-1.5em}

% \paragraph{Generalization on unseen object, unseen background, and cluttered scenes.}
\paragraph{Impact of \nbdata \  on Model Robustness}
To assess whether the data augmentation strategies in \nbdata \  enhance the robustness of VLAs,we compare RDT fine-tuned on \nbdata \ only (\nbdata) versus \nbdata \ with augmentation (\nbdata-aug).
Specifically,we select four tasks and conduct real-robot experiments under three distribution shifts: \emph{unseen objects}, \emph{unseen backgrounds}, and \emph{cluttered scenes}.
Each cell shows \emph{successes/trials} for two independent runs, as shown in \cref{tab:ablation_RDT_unseen}.
% This setup isolates whether reconstruction-driven data and simple augmentations suffice to transfer the skill to new objects, backgrounds, and cluttered environments.
% \begin{table*}[t]
% \centering
% \setlength{\tabcolsep}{0pt}
% \renewcommand{\arraystretch}{1.2}
% % \scriptsize
% \vspace{-0.7em}
% \caption{Task performance in unseen scenarios with RDT.}
% \vspace{-0.7em}
% \label{tab:ablation_RDT_unseen}
% \begin{tabularx}{0.95\textwidth}{@{}l *{6}{>{\centering\arraybackslash}X}@{}}
% \toprule
% \multirow{2}{*}{\textbf{Real-world Tasks}} 
% & \multicolumn{2}{c}{\textbf{Unseen Object}} 
% & \multicolumn{2}{c}{\textbf{Clutter Objects}} 
% & \multicolumn{2}{c}{\textbf{Unseen Background}} \\
% \cmidrule(lr){2-3} \cmidrule(lr){4-5} \cmidrule(lr){6-7}
% & \textbf{RW} & \textbf{RW-aug} 
% & \textbf{RW} & \textbf{RW-aug} 
% & \textbf{RW} & \textbf{RW-aug} \\
% \midrule
% Lift wooden cup & 4/10 & 5/10 & 4/10 & 4/10 & 0/10 & 4/10 \\
% Place milk      & 5/10 & 6/10 & 5/10 & 6/10 & 1/10 & 3/10 \\
% Upright milk    & 3/10 & 3/10 & 4/10 & 4/10 & 3/10 & 5/10 \\
% Pour cola       & 3/10 & 3/10 & 3/10 & 4/10 & 2/10 & 4/10 \\
% \midrule
% \textbf{Average} & 3.75/10 & 4.25/10(+0.5) & 4.00/10 & 4.50/10(+0.5) & 1.50/10 & 4.00/10(+2.5) \\
% \bottomrule
% \end{tabularx}
% \end{table*}
When trained on the \nbdata \ without augmentation, the fine-tuned RDT still manages to achieve some successful trials when encountering unseen objects or inferring in cluttered environments. However, when the background is altered, inducing substantial changes in the observations, the model exhibits catastrophic degradation in performance. In contrast, the fine-tuned RDT with augmented \nbdata \, shows a significant improvement in handling new observations, particularly in the unseen background setting, where the success rate increased by 25\%. These results demonstrate that training on \nbdata{}, particularly with augmentation, substantially enhances the robustness of VLA models to visual variations.

\vspace{-0.5em}

\subsection{Real-robot replay performance comparison}

\vspace{-0.3em}

% Under the same hand motion input, we retarget the hand motions to a two-finger gripper using different existing methods and execute the same set of tasks, using success rate to quantify the robustness of each mapping scheme. As evidenced by the comparisons in Tab.~\ref{tab:realrobotreplay}, our retargeting approach consistently attains higher success rates across tasks, and its definition of gripper orientation is more robust, enabling stable yet flexible task execution under diverse hand-gesture regimes.
Under the same hand joints sequence input, we retarget the motions to a two-finger gripper using both existing methods and our proposed approach as discussed in Sec.~\ref{section:crosser}, executing an identical set of tasks and measuring success rates to quantitatively evaluate the robustness of each mapping scheme. As shown in Tab.~\ref{tab:realrobotreplay}, our retargeting method consistently achieves higher success rates across all manipulation tasks. Its refined formulation of gripper orientation proves more robust, enabling stable yet flexible execution under diverse and dynamic hand-gesture regimes.

\vspace{-1.0em}
\section{Conclusion and Discussion}
% Our helical data engine \nbname \ is designed to transform real-world hand-object interaction videos into cross-domain robotic learning supervision. By leveraging high-fidelity reconstruction, multi-stage optimization, and cross-embodiment retargeting, \nbname \ offers a scalable framework for generating training data that is physically consistent and directly applicable to diverse robotic platforms. We demonstrated its effectiveness through an innovative data pipeline that includes simulation-augmented data augmentation and domain randomization, allowing for the generation of a large-scale multimodal dataset that supports various VLAs and imitation learning models.However, our real-world cross-embodiment experiments, particularly on dexterous hands and humanoid robots, remain limited in scope, and more extensive hardware validation is left for future work.

\noindent \textbf{Conclusion.} Our data engine, \nbname, transforms real-world HOI videos into cross-embodiment robotic learning supervision by integrating high-fidelity hand–object reconstruction, physically plausible trajectory optimization, and cross-embodiment retargeting. Leveraging unified HOI representation that is inherently robust to existing embodiment variations, \nbname \ provides a scaleble framework for generating robotic training data. The framework is further strengthened with a rich suite of simulation-based data augmentation strategies.
Based on two self-developed hardware platforms and existing public HOI datasets, we have constructed a large-scale HOI-to-Robot dataset \nbdata, which supports VLA models and imitation-learning approaches. 
Furthermore, our study establishes the first quantitatively validated evidence that HOI data can serve as an effective upstream modality for training models across diverse downstream, embodiment-specific tasks.

\noindent \textbf{Limitations and Future Work.} The scope of our real-world cross-embodiment experiments—particularly those involving dexterous hands and humanoid remains limited. Expanding cross-domain validation in both real-world and simulated settings will be an important direction in future.

\section*{Acknowledgement} 
This research was funded through the National Key Research and Development Program of China (Project No. 2022YFB36066), in part by the Shenzhen Science and Technology Project under Grant (KJZD20240903103210014).

% However, the scope of our real-world cross-embodiment experiments—particularly those involving dexterous hands and humanoid robots—remains limited. Expanding cross-domain validation in both real-world and simulated settings will be an important direction for future work.

% \section*{Ethics Statement}
% This research adheres to ethical guidelines in all aspects of the study.\nbname \,is a system that converts in-the-wild human hand-object interaction (HOI) videos into embodied supervision for cross-domain robotic learning. In our research, we used videos from the internet, but their usage is strictly limited to academic purposes. Any harmful use is not intended or encouraged.

% \section*{Reproducibility statement}
% To reproduce our retargeting results in both real-world and simulated environments, please consult the ~\cref{appendix:Two Mapping Algorithm}, where we present the pseudocode for the retargeting process. Regarding the real-world performance evaluation of the four VLA/IL policies, the code is fully sourced from the open-source repository, with implementation specifics provided in the ~\cref{appendix:implement details VLA/IL}.

{
    \small
    \bibliographystyle{ieeenat_fullname}
    \bibliography{main}
}

\clearpage
% \newpage

\setcounter{page}{1}
\appendix

{
\onecolumn
\centering
\section*{\fontsize{16pt}{21pt}\selectfont RoboWheel: A Data Engine from Real-World Human Demonstrations for Cross-Embodiment Robotic Learning}
\section*{Supplementary Material}

\etocdepthtag.toc{mtappendix}
\etocsettagdepth{mtchapter}{none}
\etocsettagdepth{mtappendix}{subsection}
\tableofcontents 
}

%\clearpage
%\input{sec/problem_chanllenge}
\newcolumntype{Y}{>{\raggedright\arraybackslash}X} % left-aligned, no full justification
\newcommand{\MeaningW}{0.50\textwidth}

\clearpage
\section{Notations}
\begin{table}[!h]
\centering
\small
\begin{tabularx}{\textwidth}{@{}l l@{} Y }
\toprule
\textbf{Symbol} & \textbf{Meaning} & \textbf{Notes / Space} \\
\midrule
$\{I_t\}_{t=1}^T$ & Input frame sequence & RGB / RGB-D \\
$K$ & Camera intrinsics & Known or estimated \\
%$(R_{wc},\, t_{wc})$ & World-to-camera extrinsics (rotation/translation) & $R_{wc}\!\in\!SO(3)$, $t_{wc}\!\in\!\mathbb{R}^3$ \\
$T^w_c$ & Camera pose in world frame & $SE(3)$; often $(R_{wc}, t_{wc})$ \\
$\Pi(\cdot)$ & Pinhole projection operator & 3D $\to$ pixel coordinates \\
%$\Pi_{xz}(\cdot)$ & Projection onto $xz$-plane & Removes $y$ component \\
$\mathbf{h}_t$ & Hand state at time $t$ & $(\theta_h(t),\, R^w_h(t),\, t^w_h(t))$ \\
$\theta_h(t)$ & Hand(MANO) parameters & Joint angles \\
$R^w_h(t),\, t^w_h(t)$ & Wrist pose in world (rot/trans) & $SO(3)$ and $\mathbb{R}^3$ \\
$\mathbf{p}_t$ & Object state at time $t$ & $T^w_o(t) \in SE(3)$ \\
$T^w_o(t)=(R_o(t),\, t_o(t))$ & Object pose in world (rot/trans) & $SE(3)$ \\
$M_o,\, \hat{M}_o$ & Object mesh (metric / up-to-scale) & Triangle mesh + texture \\
$s_o$ & Object scale factor & From depth–mesh alignment \\
$m_t,\, D_t$ & Object mask and depth map at $t$ & Segmentation and depth \\
$P_t,\, P$ & Back-projected points (per-frame / global) & From $D_t$ and $K$ \\
$\mathrm{AABB}(\cdot)$ & Axis-aligned bounding box & Use diagonal via $\mathrm{diag}(\cdot)$ \\
$\mathbf{A}$ & Canonical action space & Unified by C2W and facing origin \\
$T^A_w=(R^{w\rightarrow A},\, t^{w\rightarrow A})$ & World-to-action-space transform & Coordinate re-alignment \\
$\Delta^2(\cdot)$ & Second-order temporal difference & Jitter suppression \\
$\rho(\cdot)$ & Robust loss (e.g., Geman–McClure) & For reprojection errors \\
%$\mathrm{IoU}(\cdot,\cdot)$ & Intersection-over-Union & Contour/overlap consistency \\
$d(\cdot,\cdot)$ & Point–set / point–surface distance & For proximity/attraction \\
$V_h,\, V_o$ & Hand mesh vertices / object surface points & Geometry sets \\
$\phi_h(\mathbf{x};t)$ & Hand signed distance field (SDF) & Positive outside (as used here) \\
$TSDF_o$ & object truncated signed distance function & negative inside and zero else \\
$\tilde{\mathbf{q}}_i(t)$ & World coords of sampled object surface point & $R_o(t)\,\mathbf{q}^{\rm loc}_i + t_o(t)$ \\
$\mathcal{N}_t$ & Near-contact candidate set & $|\phi_h|<\tau_{\rm band}$ or Top-$K$ \\
$\|\cdot\|_{F}$ & Frobenius norm & For rotation log etc. \\
$\log_{SO(3)}(\cdot)$ & Lie-group log map on $SO(3)$ & Rotation discrepancy \\
$d_{SE(3)}(\cdot,\cdot)$ & Geodesic distance on $SE(3)$ & Pose discrepancy \\
$\mathrm{E}_\pi[\cdot]$ & Expectation under policy $\pi$ & RL objective \\
$r_t$ & Instantaneous reward & Weighted components \\
% $w_{\mathrm{stab}},w_{\mathrm{reach}},w_{\mathrm{con}},w_{\mathrm{pen}},w_{\mathrm{eng}},w_{\mathrm{goal}}$ & Reward weights & Stability/reach/contact/penetration/energy/goal \\
$\psi_{\mathrm{limits}}(\cdot)$ & Joint-limit margin reward & Prefer mid-range \\
$\pi_\theta$ & Residual policy network & Added to $a^{\mathrm{IK}}$ \\
%$a^{\mathrm{IK}}_t$ & Inverse-kinematics baseline action & ManipTrans-style residual control \\
$T_{\mathrm{rel}}(t)$ & Relative pose $T_h^{-1}(t)\,T_o(t)$ & Hand–object relative pose \\
%$\mathrm{ADD\mbox{-}S}$ & Average Distance (symmetric objects) & Pose error metric (cm) \\
$R_g,\, p_g$ & Gripper pose (rot/trans) & From hand keypoints \\
$q^{(r)}_t$ & Joint configuration of robot $r$ at $t$ & From bounded-rate IK \\
$\phi_{\mathrm{lim}}(q)$ & Joint-limit penalty & IK constraint term \\
%$m(q)$ & Manipulability & Avoid singularities \\
$S=\mathrm{diag}(-1,1,1)$ & Left–right mirror matrix & About sagittal plane \\
$R_y(\pi)$ & Rotation about $y$ by $\pi$ & Used with $S$ to set facing \\

$\varphi_{\mathrm{sem}}(\cdot)$ & Semantic embedding (text–shape) & For semantic similarity \\
$\gamma^t$ & Discount factor power & RL return discounting \\
\bottomrule
\end{tabularx}
\label{tab:symbols}
\end{table}

% $w_{\mathrm{stab}},w_{\mathrm{reach}},w_{\mathrm{con}},w_{\mathrm{pen}},w_{\mathrm{eng}},w_{\mathrm{goal}}$ & Reward weights & Stability/reach/contact/penetration/energy/goal \\

% $S(\cdot)$ & Similarity score (object replacement) & $\alpha\,\mathrm{CD}+\beta(1-\mathrm{IoU}_{\mathrm{AABB}})+\gamma\langle \varphi_{\mathrm{sem}},\cdot\rangle$ \\
% $\mathrm{CD}(\cdot,\cdot)$ & Chamfer distance (surface samples) & Shape similarity \\
% $\mathrm{IoU}_{\mathrm{AABB}}$ & IoU of AABBs & Coarse shape compatibility \\
% 
% $L_{\mathrm{jnt}}$ & 2D joint reprojection loss (hand) & Keypoint alignment \\
% $L_{\mathrm{sil}}$ & Silhouette/mask consistency loss (object) & Differentiable rasterization \\
% $L_{\mathrm{prox}}$ & Hand–object proximity/attraction loss & Encourages approach \\
% $L_{\mathrm{tmp}}$ & Temporal smoothness regularizer & On $t_o(t)$ etc. \\
% $L_{\mathrm{pen}}$ & Interpenetration penalty (SDF-based) & $\max(0,-\phi_h)^2$ \\
% $L_{\mathrm{con}}$ & Contact zero-level constraint & $\phi_h(\tilde{\mathbf{q}}_i)^2$ \\
% $L_{\mathrm{reg}}$ & Generic regularizer & Priors / bounds \\

\clearpage

\section{Transform to Canonical Action Space}
\label{canoical space}
\begin{figure*}[t]
    \centering
    % \vspace{-1.5em}
    \begin{overpic}[width=0.9\linewidth]{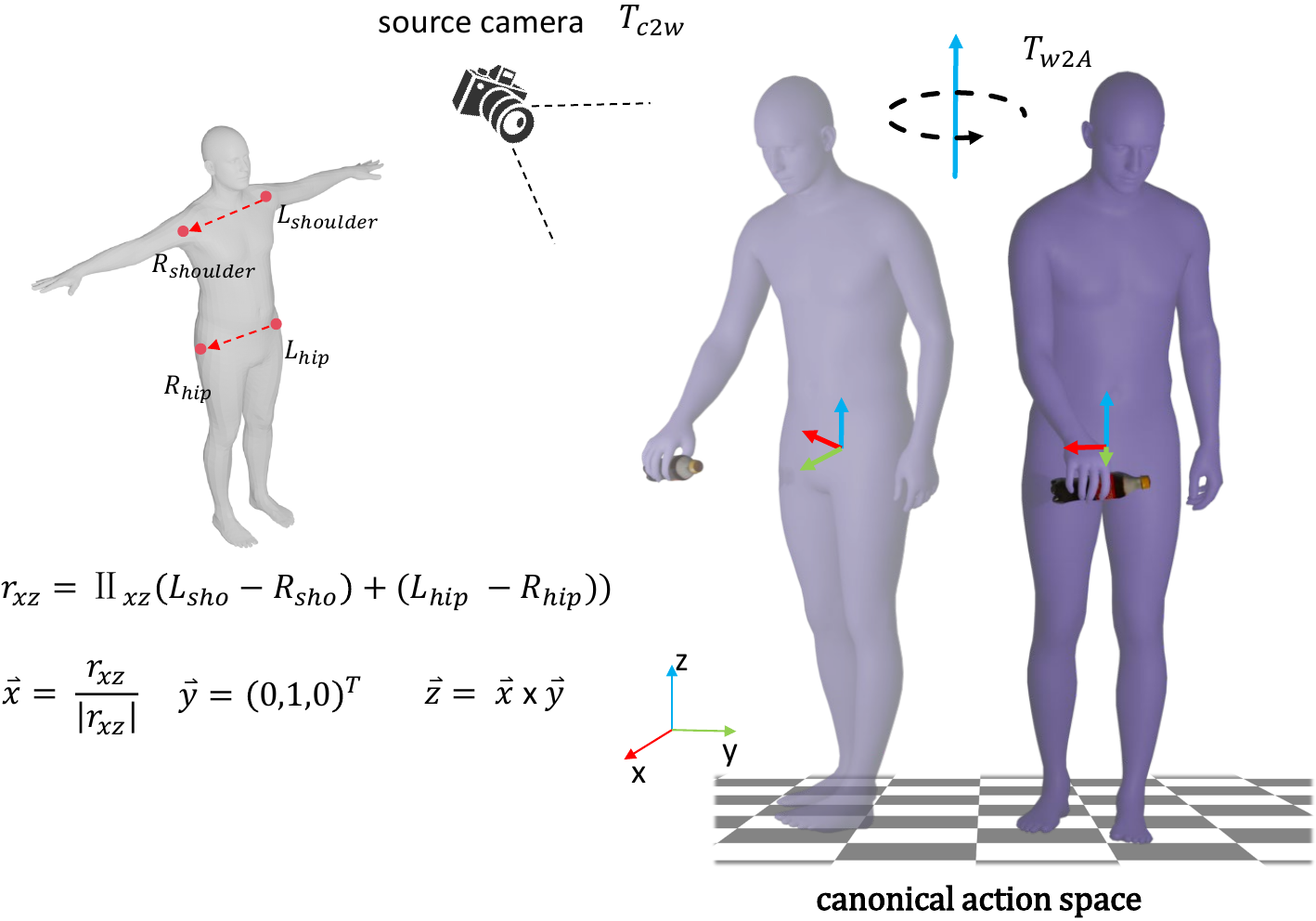}
    \end{overpic}
    \vspace{-1em}
    \caption{Our pipeline decouples the reconstructed hand-object motions from the specific camera viewpoint of the source video. We first lift the reconstructions to a consistent world frame using the camera-to-world transformation $T_{c2w}$. Subsequently, we normalize these trajectories into a canonical action space via $T_{w2\mathbf{A}}$. This two-step alignment ensures that the retargeted actions maintain a consistent approach direction and kinematic interpretation across all robotic embodiments.}
    \vspace{-1.8em}
    \label{fig:canonicalspace}
\end{figure*}

Real-world HOI videos are captured under arbitrary viewpoints, which leads to view-dependent reconstructions of both hand and object. To eliminate this inconsistency, we adopt a lifting procedure, as shown in ~\cref{fig:canonicalspace}. In the first step, all reconstructed HOI results are transformed from their respective camera coordinate systems into the same world coordinate system. In the second step, the trajectories in the world coordinate system are further normalized into a unified canonical action space, ensuring that interaction trajectories from heterogeneous sources become retargetable.

\textbf{Step 1: Estimate $(K, T_c^w)$ and lift to the world frame.}
We assume a static-camera prior and estimate camera intrinsics $K$ and a
(time-invariant) camera-to-world transform $T_c^w=(R_{wc},t_{wc})$ with
DROID-SLAM~\cite{droid}, optimizing sparse/semi-dense reprojection together with temporal smoothness:
% \begin{equation}
%     \min_{K,\,T_c^w}
%     \sum_{t,i}\!\left\|
%     \Pi\!\big(K,(T_c^w)^{-1};X_i\big)-u_{i,t}
%     \right\|_2^2 
%     +\lambda\!\left(
%     \|\Delta t_{wc}\|_2^2+\big\|\log_{\mathrm{SO}(3)}(R_{wc}^{\!\top}R_{wc}^{\,+})\big\|_2^2
%     \right),
%     \label{eq:ba-unified}
% \end{equation}
\begin{equation}
\begin{split}
    \min_{K,\,T_c^w}
    &\sum_{t,i}\!\left\|
        \Pi\!\big(K,(T_c^w)^{-1};X_i\big)-u_{i,t}
      \right\|_2^2 \\
    &
    + \lambda\!\left(
        \|\Delta t_{wc}\|_2^2
        + \big\|\log_{\mathrm{SO}(3)}(R_{wc}^{\!\top} R_{wc}^{\,+})\big\|_2^2
      \right),
\end{split}
\label{eq:ba-unified}
\end{equation}
where $(\cdot)^{+}$ denotes the next keyframe. We adopt the keyframe solution as the clipwise $T_c^w$. We also evaluated DPVO and COLMAP but found DROID-SLAM more stable on our setting. With the fixed $T_c^w$ for each clip, we could project our estimated $\mathbf{h}_t, \mathbf{p}_t$ convert to the world frame:
% \begin{equation}
%     T_h^w(t)=T_c^w\circ T_h^c(t),\qquad
%     T_o^w(t)=T_c^w\circ T_o^c(t).
%     \label{eq:cam2world}
% \end{equation}

% Step 2 world to face
% \textbf{Step 2: Align to canonical action space $\mathcal{A}$.}
% Concretely, with left/right hip and shoulder positions
% $p_{\mathrm{hip}}^{L/R},\,p_{\mathrm{sho}}^{L/R}$ in $\{\mathcal{Z}\}$, we form a lateral vector on the $xz$-plane,
% \[
% v_{\mathrm{lat}} \;=\; \Pi_{xz}\!\big((p_{\mathrm{hip}}^{L}-p_{\mathrm{hip}}^{R}) + (p_{\mathrm{sho}}^{L}-p_{\mathrm{sho}}^{R})\big),
% \]

% We specify a standard operational frame with $z$-up, a fixed facing direction on $y$, and
% $x=y\times z$. Concretely, $y_{\mathcal{A}}$ is set by the dominant interaction direction
% (e.g., average hand$\to$object approach vector), $z_{\mathcal{A}}$ follows scene up (gravity /
% ground plane), and $x_{\mathcal{A}}$ is completed by the right-hand rule. After orthonormalization,
% these axes define $R_{w\!\to\!\mathcal{A}}\in\mathrm{SO}(3)$. We center the action at the object
% (by default the first salient frame $t_0$) via
% $t_{w\!\to\!\mathcal{A}}=-R_{w\!\to\!\mathcal{A}}\,t_o^w(t_0)$, giving
% $T_w^{\mathcal{A}}=(R_{w\!\to\!\mathcal{A}},t_{w\!\to\!\mathcal{A}})$.

\textbf{Step 2: Align to canonical action space $\mathcal{A}$.}  
Given the left/right hip and shoulder positions 
$p_{\mathrm{hip}}^{L/R},\,p_{\mathrm{sho}}^{L/R}$ 
in the world coordinate system $\{\mathcal{Z}\}$, we first compute a lateral reference vector on the $xz$-plane:
\[
v_{\mathrm{lat}} \;=\; \Pi_{xz}\!\big((p_{\mathrm{hip}}^{L}-p_{\mathrm{hip}}^{R}) + (p_{\mathrm{sho}}^{L}-p_{\mathrm{sho}}^{R})\big).
\]

We then construct the canonical frame $\mathcal{A}$ with the following conventions:
\begin{itemize}
    \item $z_{\mathcal{A}}$ is aligned with the scene up direction (gravity / ground normal).
    \item $y_{\mathcal{A}}$ is determined by the dominant interaction direction (e.g., the average hand$\to$object approach vector).
    \item $x_{\mathcal{A}}$ is obtained by the right-hand rule, ensuring orthonormality.
\end{itemize}

After orthonormalization, these axes form the rotation matrix 
$R_{w\!\to\!\mathcal{A}} \in \mathrm{SO}(3)$.  
Finally, we center the action trajectory at the object position (by default, at the first salient frame $t_0$), giving the translation 
$t_{w\!\to\!\mathcal{A}} = -\,R_{w\!\to\!\mathcal{A}}\,t_o^w(t_0)$ 
and the resulting rigid transformation 
$T_w^{\mathcal{A}} = (R_{w\!\to\!\mathcal{A}},\, t_{w\!\to\!\mathcal{A}})$.

%\clearpage
\section{Hand Gesture to Robot Arm}
\subsection{Mapping Algorithm}
\label{appendix:Two Mapping Algorithm}
For the two different gesture types, whole-hand and finger-only, we designed corresponding mapping schemes to translate hand motions to a two-fingered gripper. For whole-hand gestures, our method primarily leverages 3d joints on the palm plane to define the gripper's orientation and spatial position. For finger-only gestures, we incorporate fingertip positions to accommodate dexterous manipulation. The detailed mapping algorithm pseudocode under two gestures is shown in Fig.~\ref{fig:mapping pesudocode}.

\begin{figure*}[!h]
\centering

\begin{minipage}[t]{0.48\textwidth}
\begin{algorithm}[H]
\caption{Whole-Hand (Palm-Involved) Gripper Pose Construction}

\textbf{Input:} wrist $\mathbf{k}_{\mathrm{wri}}$, 
index MCP $\mathbf{k}_{\mathrm{ind}}^{\mathrm{mcp}}$, \\
ring MCP $\mathbf{k}_{\mathrm{ring}}^{\mathrm{mcp}}$ \\ [0.5em]
% , chirality $\diamond\!\in\!\{L,R\}$, offset $d_z$.\\ [0.5em]
\textbf{Output:}~Gripper~pose $(R_g,\mathbf{p}_g)$.\\[-0.5em]
\begin{algorithmic}
\STATE $\mathbf{w}\!\leftarrow\!\mathbf{k}_{\mathrm{wri}}$, $\mathbf{i}\!\leftarrow\!\mathbf{k}_{\mathrm{ind}}^{\mathrm{mcp}}$, $\mathbf{r}\!\leftarrow\!\mathbf{k}_{\mathrm{ring}}^{\mathrm{mcp}}$

 \hspace*{1.5em}\textit{// Extract keypoints} 
\STATE $\mathbf{o}\leftarrow (\mathbf{w}+\mathbf{i}+\mathbf{r})/3$ \hspace*{1.0em}\textit{// palm origin}
% \STATE \hspace*{1.5em}\textit{// palm origin}
\STATE $v_x \leftarrow (\mathbf{r}-\mathbf{w})$\hspace*{1.0em}\textit{// X-axis direction }
% \STATE \hspace*{1.5em}\textit{// X-axis direction }
\STATE $\bar{\mathbf{x}} \leftarrow v_x/(\textsc{NORMALIZE}(v_x)+10^{-8})$
\STATE \hspace*{1.5em}\textit{// Normalized X-axis}
\STATE $v_z \leftarrow \textsc{CROSS\_PRODUCT}(\mathbf{i}-\mathbf{w},\ \mathbf{r}-\mathbf{w})$
\STATE \hspace*{1.5em}\textit{// Z-axis (palm normal)}
\STATE $\bar{\mathbf{z}} \leftarrow v_z/(\textsc{NORMALIZE}(v_z)+10^{-8})$
\STATE \hspace*{1.5em}\textit{// Normalized Z-axis}

\STATE $\bar{\mathbf{y}} \leftarrow \textsc{CROSS\_PRODUCT}(\bar{\mathbf{z}},\bar{\mathbf{x}})$
\STATE \hspace*{1.5em}\textit{// Y-axis direction}
\STATE $\bar{\mathbf{z}} \leftarrow \textsc{SIGN}(\diamond)\cdot \bar{\mathbf{z}}$
%\textit{// chirality: $\textsc{SIGN}(R)\!=\!+1$, $\textsc{SIGN}(L)\!=\!-1$}
\STATE $R_g \leftarrow \textsc{CONCATENATE}([\bar{\mathbf{x}},\bar{\mathbf{y}},\bar{\mathbf{z}}])$
\STATE \hspace*{1.5em}\textit{// Rotation matrix}
% \STATE $\mathbf{p}_g \leftarrow \mathbf{o} + d_z\,\bar{\mathbf{z}}$\hspace*{0.5em}\textit{// Position }
\STATE$\mathbf{p}_g \leftarrow \mathbf{o} + d_z\,\bar{\mathbf{z}}$\hspace*{0.5em} \textit{// Position } \;
\end{algorithmic}
% \STATE \textbf{return} $(R_g,\mathbf{p}_g)$
\textbf{Return}~$(R_g,\mathbf{p}_g)$
\end{algorithm}

\end{minipage}
\hfill
\begin{minipage}[t]{0.48\textwidth}
\begin{algorithm}[H]
\caption{Finger-Only (Pinch/Precision) Gripper Pose Construction}

\textbf{Input:} index TIP $\mathbf{k}_{\mathrm{ind}}^{\mathrm{tip}}$, index MCP $\mathbf{k}_{\mathrm{index}}^{\mathrm{mcp}}$ , thumb tip $\mathbf{k}_{\mathrm{thumb}}^{\mathrm{tip}}$, thumb MCP $\mathbf{k}_{\mathrm{thumb}}^{\mathrm{mcp}}$ \\ [0.5em]
% , chirality $\diamond\!\in\!\{L,R\}$, offset $d_z$.\\ [0.5em]
\textbf{Output:}~Gripper~pose $(R_g,\mathbf{p}_g)$.\\[-0.5em]
\begin{algorithmic}
\STATE $\mathbf{i}\!\leftarrow\!\mathbf{k}_{\mathrm{ind}}^{\mathrm{tip}}$, 
$\mathbf{m}\!\leftarrow\!\mathbf{k}_{\mathrm{ind}}^{\mathrm{mcp}}$, 
$\mathbf{t}\!\leftarrow\!\mathbf{k}_{\mathrm{thumb}}^{\mathrm{tip}}$, $\mathbf{r}\!\leftarrow\!\mathbf{k}_{\mathrm{thumb}}^{\mathrm{mcp}}$

\hspace*{1.5em}\textit{// Extract keypoints} 

\STATE $\mathbf{o}\leftarrow (\mathbf{t}+\mathbf{i})/2$\hspace*{1.0em}\textit{// palm origin}
% \STATE \hspace*{1.5em}\textit{// palm origin}
\STATE $v_z \leftarrow (\mathbf{i}-\mathbf{m})$\hspace*{1.0em}\textit{// Z-axis direction }
% \STATE \hspace*{1.5em}\textit{// Z-axis direction }
\STATE $\bar{\mathbf{z}} \leftarrow v_z/(\textsc{NORMALIZE}(v_z)+10^{-8})$
\STATE \hspace*{1.5em}\textit{// Normalized Z-axis}
\STATE $v_y \leftarrow \textsc{CROSS\_PRODUCT}(\mathbf{i}-\mathbf{m},\ \mathbf{m}-\mathbf{r})$

\hspace*{1.5em}\textit{// Y-axis (palm normal)}
\STATE $\bar{\mathbf{y}} \leftarrow v_y/(\textsc{NORMALIZE}(v_y)+10^{-8})$
\STATE \hspace*{1.5em}\textit{// Normalized Y-axis}

\STATE $\bar{\mathbf{x}} \leftarrow \textsc{CROSS\_PRODUCT}(\bar{\mathbf{y}},\bar{\mathbf{z}})$
\STATE \hspace*{1.5em}\textit{// X-axis direction}
\STATE $\bar{\mathbf{z}} \leftarrow \textsc{SIGN}(\diamond)\cdot \bar{\mathbf{z}}$
%\textit{// chirality: $\textsc{SIGN}(R)\!=\!+1$, $\textsc{SIGN}(L)\!=\!-1$}
\STATE $R_g \leftarrow \textsc{CONCATENATE}([\bar{\mathbf{x}},\bar{\mathbf{y}},\bar{\mathbf{z}}])$
\STATE \hspace*{1.5em}\textit{// Rotation matrix}
\STATE $\mathbf{p}_g \leftarrow \mathbf{o} $
\hspace*{0.5em}\textit{// Position}
\end{algorithmic}
% \STATE \textbf{return} $(R_g,\mathbf{p}_g)$
\textbf{Return}~$(R_g,\mathbf{p}_g)$
\end{algorithm}

\end{minipage}
\caption{Following our gesture classification scheme, we divide grasping into two categories and combine them into the final two-finger gripper pose based on the hand joint space orientation.}
\label{fig:mapping pesudocode}
\end{figure*}

\subsection{Replay Comparison with Different Methods}
\begin{figure*}[h]
    \centering
    \begin{overpic}[width=\textwidth]{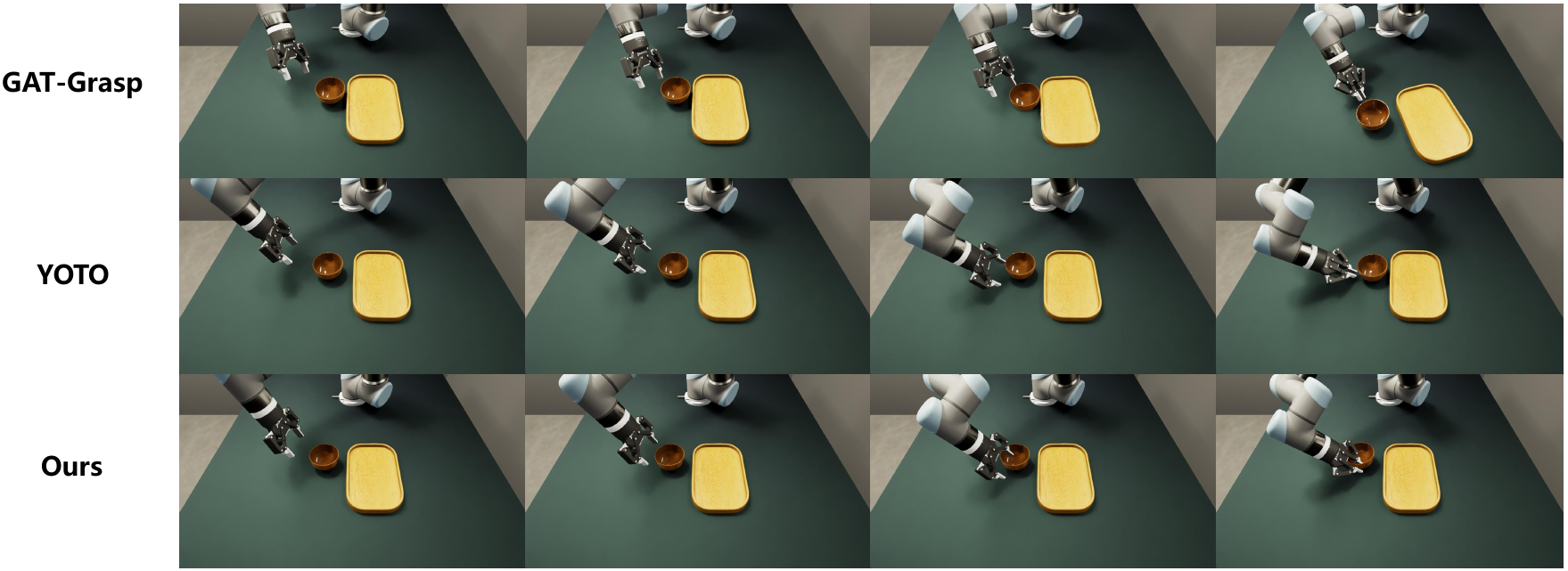}
    \end{overpic}
    \vspace{-2em}
    \caption{Replay Comparison by Different Methods: We compared the mapping method of \nbname \, with YOTO and GAT-Grasp. YOTO and GAT-Grasp result in discrepancies in gripper position or orientation mapping, leading to failure, while \nbname\, provides more accurate and reasonable mapping.}
    \label{fig:replay_comprision}
\end{figure*}

Replay Comparison by Different Methods: We compared the mapping method of \nbname \, with YOTO and GAT-Grasp. YOTO and GAT-Grasp result in discrepancies in gripper position or orientation mapping, leading to failure, while \nbname\, provides more accurate and reasonable mapping,as shown in Fig.~\ref{fig:replay_comprision}.

\subsection{Cross-Arm Retargeting Performance}
In Tab.~\ref{tab:realrobotreplay}, we report the real-world SR of different mapping methods on UR5. To further assess the generalization and scalability of our approach, we deploy the same mapping algorithm across multiple manipulators (UR5, Gen3, iiwa7, Sawyer, and Franka) in simulation. As shown in Tab.~\ref{tab:scaled_arm_success}, our method achieves consistently high performance on all arms, with SR remaining stable despite differences in kinematics and embodiments. This indicates that the proposed mapping is robust and scalable for cross-embodiment transfer.

\begin{table*}[t]
\centering
\vspace{1em}
\begin{tabular}{lccccc}
\toprule
\textbf{Task} & \textbf{UR5} & \textbf{Gen3} & \textbf{iiwa7} & \textbf{Sawyer} & \textbf{Franka} \\
\midrule
flip\_milk  & $10/10$ (100\%) & $10/10$ (100\%) & $7/10$ (70\%) & $10/10$ (100\%) & $7/10$ (70\%) \\
place\_milk & $10/10$ (100\%) & $10/10$ (100\%) & $10/10$ (100\%) & $10/10$ (100\%) & $10/10$ (100\%) \\
pour\_cola  & $10/10$ (100\%) & $10/10$ (100\%) & $10/10$ (100\%) & $10/10$ (100\%) & $10/10$ (100\%) \\
\bottomrule
\end{tabular}
\caption{Scaled success rates for 10 trials per task using the same retarget method.}
\label{tab:scaled_arm_success}
\end{table*}

\clearpage

\section{Experiment Details}
We used the \nbdata \ to train four VLA/IL policies, namely ACT, DP, RDT-1B, and Pi0, to validate the effectiveness of our data. Before training these models, we preprocessed the data into the format of observations, actions, and instructions (if required) for model training.

% \subsection{Implementation and Hyper-Parameters of 4 VLA/IL policies}
% \label{appendix:implement details VLA/IL}
% For ACT ,we trained each task for 20,000 iterations, with 90\% of the data used for training and the remaining 10\% for validation. When training DP, we kept the same training steps, learning rate, and chunk size as ACT. The specific parameter values are listed in table \ref{tab:Hyperparameters of training ACT and DP}.
% \begin{table}[h]
% \centering
% \begin{tabular}{l|c|l|c}
% \toprule
% \textbf{Hyperparameter (ACT)} & \textbf{Value} & \textbf{Hyperparameter (DP)} & \textbf{Value} \\
% \midrule
% Chunk\_Size & 16 & Chunk\_Size & 16 \\
% Hidden\_Dim & 512 & Action\_Horizon & 8 \\
% Batch\_Size & 16 & Batch\_Size & 16 \\
% Learning\_Rate & 1e-5 & Learning\_Rate & 1e-5 \\
% Dim\_Feedforward & 3200 & Observation\_Horizon & 8 \\
% Num\_Steps & 20000 & Num\_Steps & 20000 \\
% \bottomrule
% \end{tabular}
% \caption{Hyperparameters of training ACT and DP}
% \label{tab:Hyperparameters of training ACT and DP}
% \end{table}
\subsection{Implementation and Hyper-Parameters of VLA/IL policies}
\label{appendix:implement details VLA/IL}
For ACT, we trained each task for 20,000 iterations, with 90\% of the data used for training and the remaining 10\% for validation. 
When training DP, we kept the same training steps, learning rate, and chunk size as ACT. 
The specific hyperparameter values are listed in Tab.~\ref{tab:act-dp-hparams}.

\begin{table}[h]
\centering
\setlength{\tabcolsep}{4pt}      % 缩小列间距
\small                           % 整体字体变小
\begin{tabular}{l c l c}
\toprule
\multicolumn{2}{c}{\textbf{ACT}} & \multicolumn{2}{c}{\textbf{DP}} \\
\cmidrule(r){1-2}\cmidrule(l){3-4}
\textbf{Hyperparameter} & \textbf{Value} & \textbf{Hyperparameter} & \textbf{Value} \\
\midrule
Chunk size        & 16    & Chunk size          & 16    \\
Hidden dim        & 512   & Action horizon      & 8     \\
Batch size        & 16    & Batch size          & 16    \\
Learning rate     & 1e-5  & Learning rate       & 1e-5  \\
Dim feedforward   & 3200  & Observation horizon & 8     \\
Training steps    & 20000 & Training steps      & 20000 \\
\bottomrule
\end{tabular}
\caption{Hyperparameters used to train ACT and DP.}
\label{tab:act-dp-hparams}
\end{table}

\textbf{RDT} was pretrained for 100,000 steps with a batch size of 8 per GPU on 4 GPUs, and all single-task fine-tuning was conducted for 10,000 steps with a batch size of 8 per GPU on a single GPUs. \textbf{Pi0}  was pretrained for 100,000 steps with a batch size of 32 on 8 GPUs, and all fine-tuning was performed for 30,000 steps using the same batch size on a single GPU.

% The other hyperparameters for RDT and Pi0 were kept consistent with the official documentation.
All remaining hyperparameters for RDT and Pi0 were set according to their official documentation.

\subsection{Training \nbdata \ on Maniptrans}
\begin{figure*}[!h]
    \centering
    \begin{overpic}[width=0.95\linewidth]{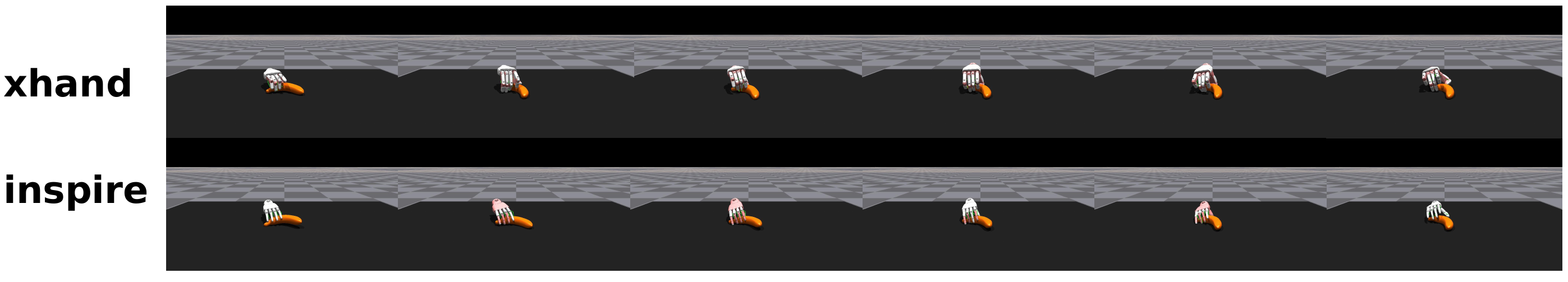}
    \end{overpic}
    
    \caption{Visualization results of maniptrans}
    \label{fig:maniptrans}
\end{figure*}

We mapped the hand gestures onto the dexterous hands and completed the training in a simulation environment. The training outcomes are shown in Fig.~\ref{fig:maniptrans}.

\subsection{Real-world Validation Results}
\begin{figure*}[!h]
    \centering
    \begin{overpic}[width=0.95\linewidth]{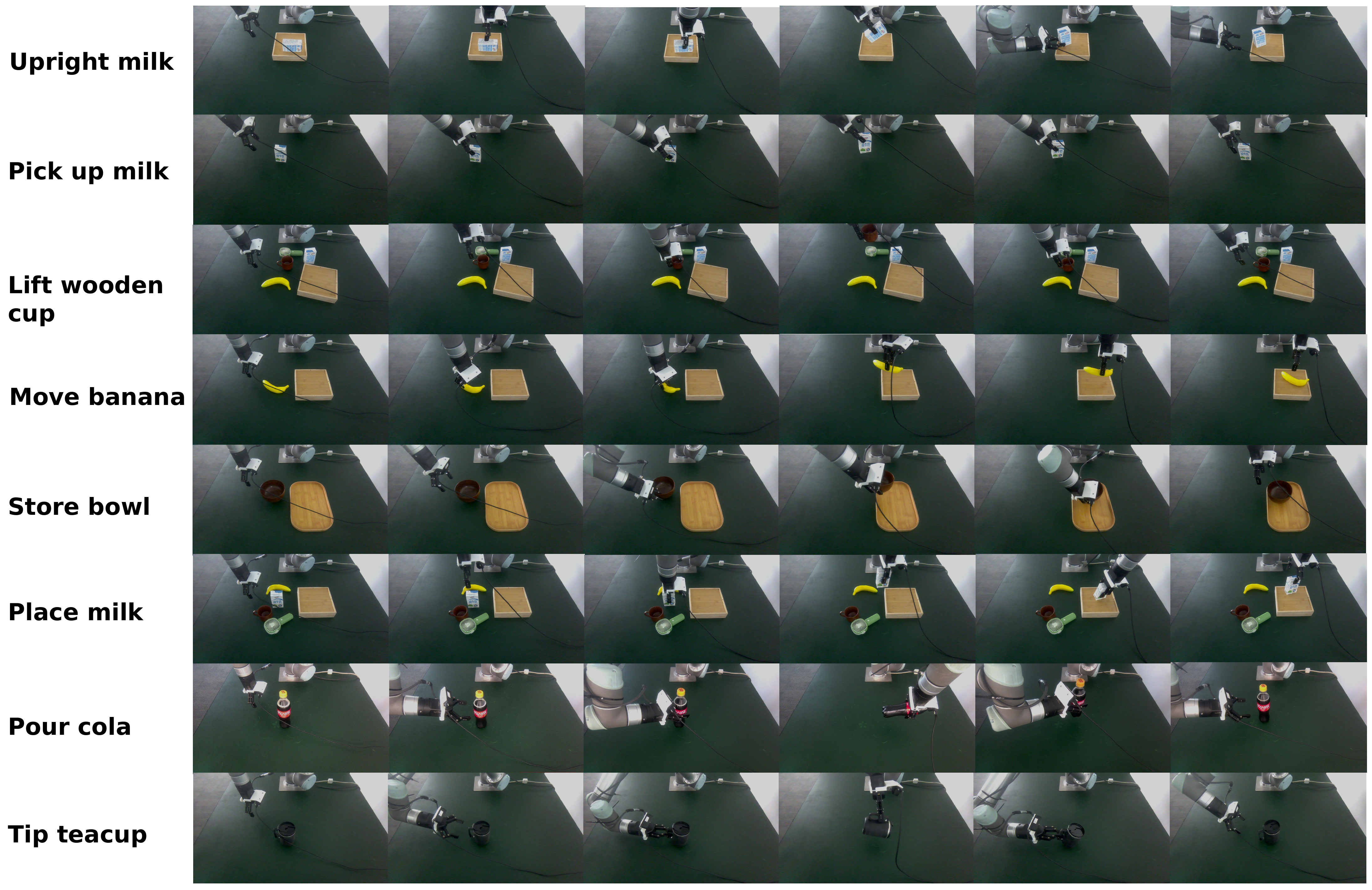}
    \end{overpic}
    \caption{Visualization of eight tasks on real robot}
    \label{fig:real robot appendix}
\end{figure*}

Here, we present the experimental results for all designed tasks conducted on physical robot(UR5),as shown in Fig.~\ref{fig:real robot appendix}.

%\section{LLM Usage}
% We utilize large language models (LLMs) to polish our articles and correct errors, including grammatical mistakes and imprecise phrasing.
\clearpage
\section{Dataset Details}
\label{sec:dataset_tasks_multidim}

We construct a large-scale dataset (\nbdata) with three subsets: 
\textit{HORA-Mocap} (multi-view mocap with tactile gloves), 
\textit{HORA-Recordings} (in-house RGB(D) recordings without tactile), 
and \textit{HORA-Public} (retargeted public HOI datasets).

\subsection{\nbdata \ Mocap Handware Setup}
\label{append:mocap}
\begin{figure*}[!h]
    \centering
    \begin{overpic}[width=\textwidth]{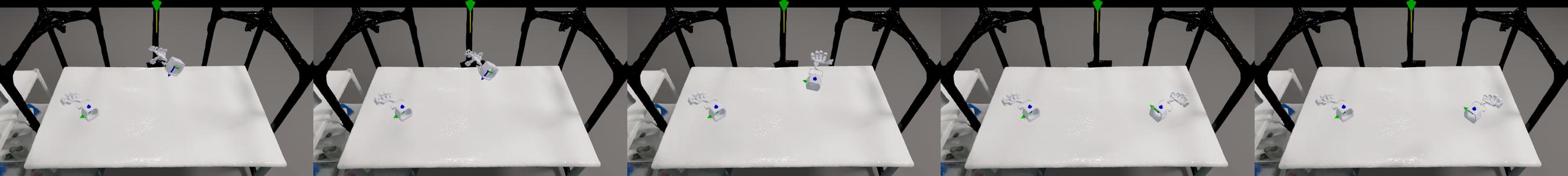}
    \end{overpic}
    \vspace{-2em}
    \caption{Visualization of the Mocap Hardware setup and gloves in simulation.}
    \label{fig:vis of mocap}
\end{figure*}

\paragraph{Glove}
The glove is instrumented with 16 Gen3 tactile sensors and 29 magnetic encoders. Worn on a single hand, it enables the acquisition of high-frequency tactile data. The tactile sensors are capable of detecting pressure, force, and vibration, while the magnetic encoders are used to capture precise joint angles and movements of the fingers. This combination allows for detailed hand-object interaction data to be gathered with high temporal resolution. Visualization of the Mocap Hardware setup and gloves in simulation is shown as Fig. ~\ref{fig:vis of mocap}.

\paragraph{RGBD Cameras}
Three Intel RealSense D455 RGBD cameras are used to capture depth and RGB data simultaneously. The cameras are mounted at strategic locations to ensure optimal coverage and accurate 3D spatial data. The depth cameras provide high-resolution depth maps, while the RGB cameras offer high-quality color images. For synchronized data acquisition, a synchronization cable is employed, connecting three cameras to ensure precise temporal alignment across all devices.

\paragraph{RGB Cameras}
A total of eight high-resolution RGB cameras are used for detailed visual tracking. These cameras are positioned to cover different angles, enabling comprehensive capture of the environment and subjects. The cameras are used to provide complementary visual data to the depth information provided by the RGBD cameras.

\subsection{From Glove Fitting to MANO Parameters}
\label{subsec:glove_to_mano}

We retarget glove-based kinematic fits to the MANO hand model through a unified differentiable optimization. 
Both the glove and MANO use differentiable forward kinematics (FK), allowing a smooth transition that preserves global pose and local contact geometry.

\paragraph{Glove-domain fitting.}
We first calibrate the glove using multi-view constraints. 
To ensure consistent correspondence, we define: 
(i) a link-level mapping between glove joints and MANO joints, and 
(ii) tactile sensor points assigned to glove links. 
Contact constraints are activated only for links with nonzero force.
If a link has no force, its position/normal constraints are disabled.
For links under force, we apply force-aware weights:
strong-force sensor points have weight $1.0$, while other activated points have weight $0.3$.
The glove fitting objective is
\begin{equation}
\mathcal{L}_{\text{glove}}=
\mathcal{L}_{\text{contact}}+
\mathcal{L}_{\text{wrist}}+
\mathcal{L}_{\text{jlim}}+
\mathcal{L}_{\text{smooth}}+
\mathcal{L}_{\text{anat}},
\end{equation}
where $\mathcal{L}_{\text{contact}}$ aligns positions and normals at tactile points, 
$\mathcal{L}_{\text{wrist}}$ constrains the wrist pose, 
$\mathcal{L}_{\text{jlim}}$ penalizes joint-limit violations, 
$\mathcal{L}_{\text{smooth}}$ enforces temporal smoothness, 
and $\mathcal{L}_{\text{anat}}$ regularizes anatomical plausibility. 
All frames are optimized in parallel. 
To handle wrist floating, we prepend a free 6-DoF transform before the glove wrist base.

\paragraph{Retargeting initialization.}
The optimized glove motion is used to initialize MANO. 
We copy the glove global wrist transform (including the 6-DoF floating pose) to MANO's root pose,
and map each glove joint rotation to its corresponding MANO joint using the predefined correspondence.
This yields an initial MANO pose close to the contact-consistent glove fit.

\paragraph{MANO refinement.}
Starting from the retargeted initialization, we refine MANO parameters under the same constraints, 
now expressed on MANO joints/vertices.
MANO pose is parameterized by 16 articulated joints, each with 3-DoF axis-angle rotations, i.e.,
\begin{equation}
\boldsymbol{\theta}\in\mathbb{R}^{16\times 3},
\end{equation}
together with a global wrist pose.
We minimize
\begin{equation}
\mathcal{L}_{\text{mano}}=
\mathcal{L}_{\text{contact}}+
\mathcal{L}_{\text{wrist}}+
\mathcal{L}_{\text{jlim}}+
\mathcal{L}_{\text{smooth}}+
\mathcal{L}_{\text{anat}},
\end{equation}
using the same force-aware contact weights transferred from glove links.
This refinement produces anatomically valid MANO poses while preserving tactile interaction geometry.

\subsection{Task Description for \nbdata}
\begin{figure*}[t]
    \centering
    % \vspace{-1.5em}
    \begin{overpic}[width=0.9\linewidth]{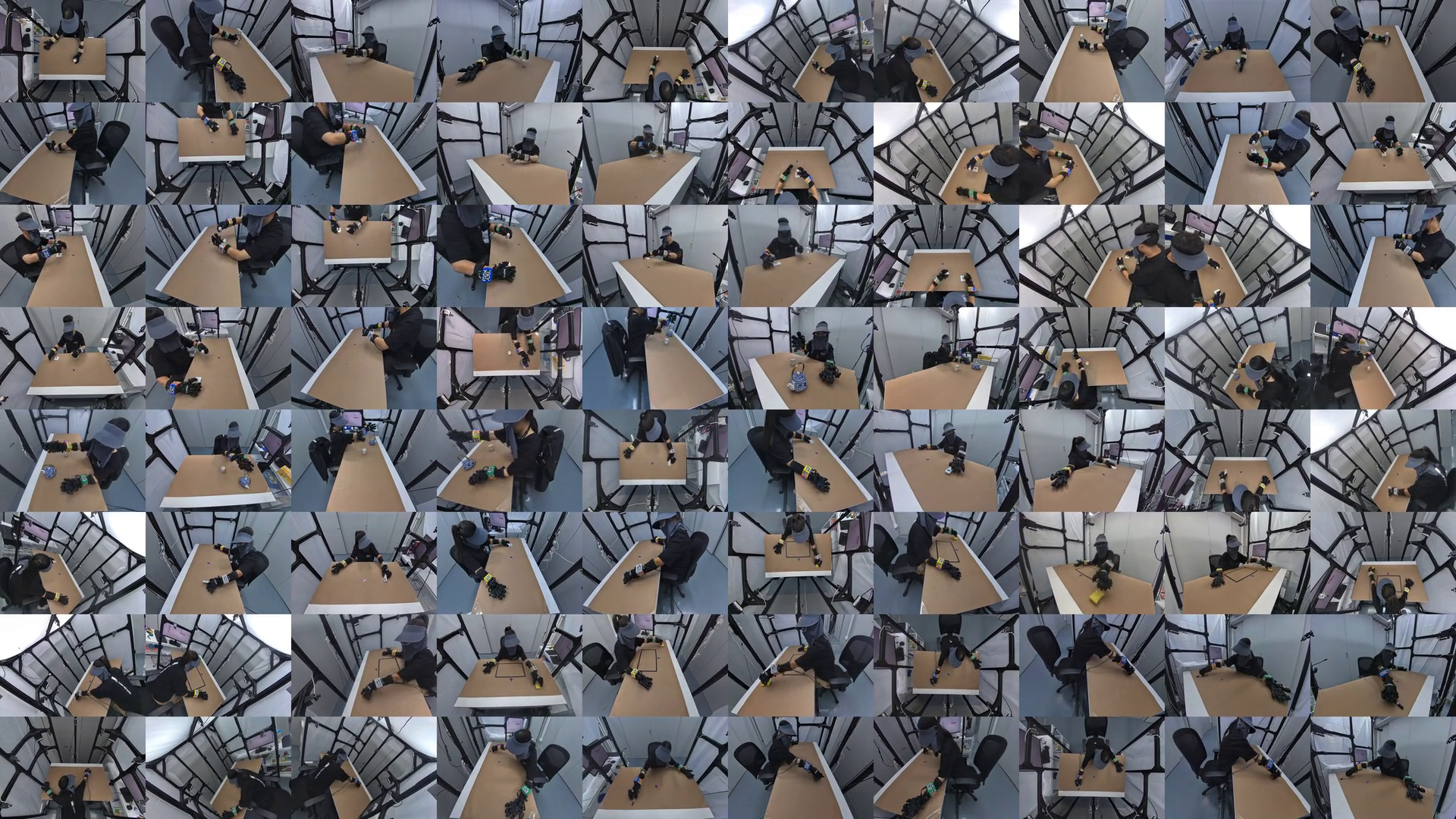}
    \end{overpic}
    %\vspace{-1em}
    \caption{\nbdata-Mocap Overview,}
    \label{fig:paxini_view}
\end{figure*}

The \nbdata-Mocap subset is designed to provide a compact yet diverse set of everyday household manipulation skills captured with high-fidelity motion tracking. We curate tasks that range from simple, atomic primitives (e.g., pick-and-place, pressing, pouring, inserting, and opening/closing) to mid- and long-horizon activities that require multi-step coordination and object-centric reasoning (e.g., pouring water into a cup, storing blocks, tabletop cleaning, tong-based picking, and bulb installation). To reflect realistic human hand usage, the task taxonomy is simplified into one-handed and two-handed categories, covering both single-handed interactions and cooperative bimanual behaviors such as rotating a cap, folding clothes, scanning items, and table setting. For detailed task descriptions and visualization, please refer to the Tab.\ref{tab:hora_mocap_tasks_en_simple} and Fig.\ref{fig:paxini_view}. The resulting task suite offers a structured benchmark for learning and evaluating dexterous manipulation policies under varied object types, action primitives, and temporal horizons.

\begin{table}[h]
\centering
\scriptsize
\setlength{\tabcolsep}{3pt}
\renewcommand{\arraystretch}{1.15}
\begin{tabular}{c p{0.16\linewidth} p{0.18\linewidth} p{0.30\linewidth} p{0.30\linewidth}}
\toprule
\textbf{\#} & \textbf{Task type} & \textbf{Primitive / skill} & \textbf{Manipulated objects} & \textbf{Task description} \\
\midrule
1  & One-handed & pick\&place & Coconut water bottle (350\,ml) & Pick up and place the object. \\
2  & One-handed & pick\&place & Black marker & Pick up and place the object. \\
3  & One-handed & pick\&place & Lidded ceramic mug (white) & Pick up and place the object. \\
4  & One-handed & pick\&place & Claw hammer & Pick up and place the object. \\
5  & One-handed & press & Remote control & Press buttons on the remote control. \\
6  & One-handed & pour & Kettle & Pick up the kettle and perform a pouring motion. \\
7  & One-handed & insert & Gel pen + pen holder & Insert the pen into the pen holder. \\
8  & One-handed & open--close & Tea gift box & Open the tea gift box and then close it. \\
9  & One-handed & pouring water & Kettle + cup &
Place the cup at a fixed position, pick up the kettle to pour water, put down the kettle, then deliver the cup to a target position. \\
10 & One-handed & block storage & Blocks + block box &
Store blocks of different shapes into the block box. \\
11 & One-handed & tabletop wiping & Cleaning sponge & Hold the sponge and wipe the tabletop. \\
12 & One-handed & vacuum cleaning & Handheld vacuum & Use the handheld vacuum to clean the tabletop. \\
13 & One-handed & tong picking & Baking tongs + bun-shaped object &
Use tongs to pick up the bun-shaped object. \\
14 & One-handed & bulb installation & Light bulb + bulb socket &
Screw the light bulb into the socket. \\
\midrule
15 & Two-handed & pick\&place & Large plate & Pick up and place the object with both hands. \\
16 & Two-handed & pick\&place & Large juice bottle & Pick up and place the object with both hands. \\
17 & Two-handed & rotate & Bottle with cap &
Cooperatively unscrew the cap with both hands, then re-tighten it. \\
18 & Two-handed & clothes folding & T-shirt &
Fold the T-shirt and place it at a fixed position. \\
19 & Two-handed & item scanning & Scanner gun + items to scan &
Pick up an item and scan it with the scanner gun. \\
20 & Two-handed & table setting & Tableware set &
Set the table following Western-style table-setting rules. \\
\bottomrule
\end{tabular}
\vspace{-1mm}
\caption{\nbdata-Mocap subset task list.}
\label{tab:hora_mocap_tasks_en_simple}
\end{table}

\clearpage

\section{Data Augmentation}

In this section, we present various data augmentation strategies and additional details applied to more tasks.
\subsection{Cross-Embodiment}

\begin{figure*}[!h]
    \centering
    % \vspace{-1.5em}
    \begin{overpic}[width=\linewidth]{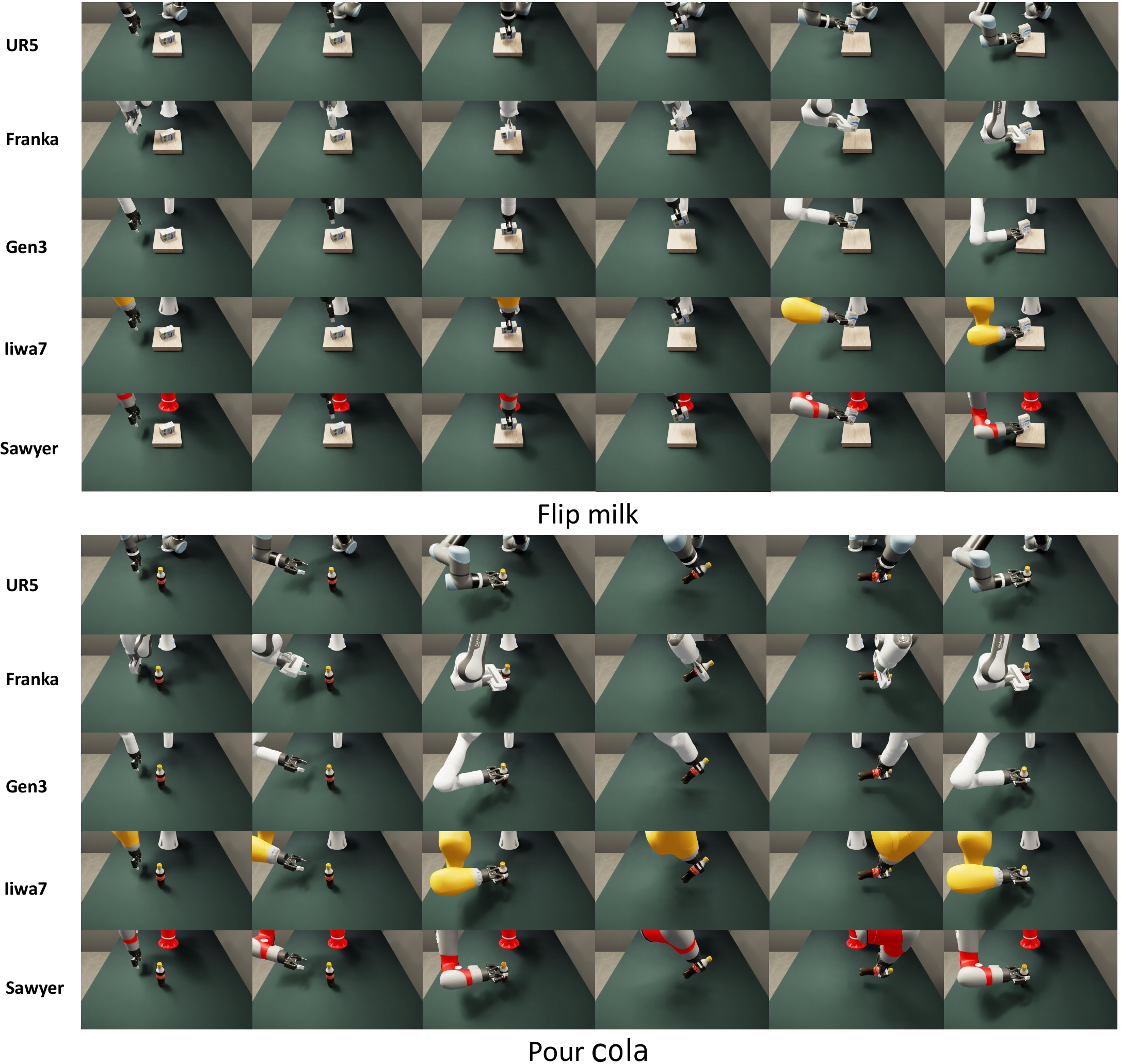}
    \end{overpic}
    \vspace{-1em}
    \caption{Visualization of robot arm augmentation:flip milk and pour Water}
    \vspace{-1.8em}
    \label{fig:append_armaug1}
\end{figure*}

\begin{figure*}[!h]
    \centering
    % \vspace{-1.5em}
    \begin{overpic}[width=\linewidth]{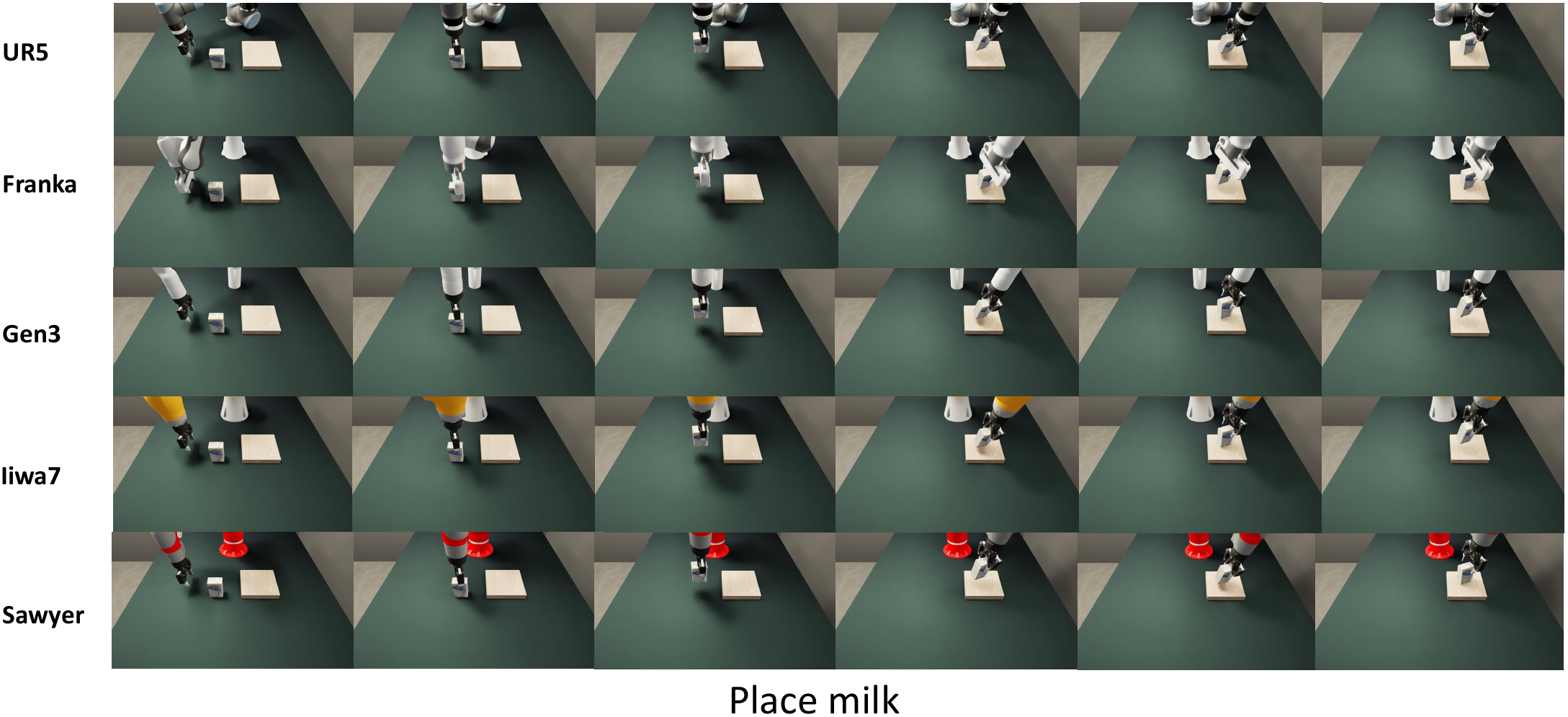}
    \end{overpic}
    \caption{Visualization of robot arm augmentation:place milk}
    \label{fig:append_armaug2}
\end{figure*}

Here, we present additional results from the data augmentation module, demonstrating motion retargeting for HOI reconstruction to different robotic arms. Fig.~\ref{fig:append_armaug1} and Fig.~\ref{fig:append_armaug2} show the retargeted motions for the tasks "flip milk," "pour water," and "place milk" to the \emph{UR5/UR5e}, \emph{Franka Emika Panda}, \emph{KUKA LBR iiwa~7}, \emph{Kinova Gen3}, and \emph{Rethink Robotics Sawyer} arms, respectively.


\vspace{1.8em}

% \begin{figure*}[!h]
%     \centering
%     % \vspace{-1.5em}
%     \begin{overpic}[width=\linewidth]{img/append_robot_arm_aug2.pdf}
%     \end{overpic}
%     \vspace{-1em}
%     \caption{}
%     \vspace{-1.8em}
%     \label{fig:pipeline}
% \end{figure*}

\subsection{Background Variation}
\begin{figure*}[!h]
    \centering
    \begin{overpic}[width=0.95\linewidth]{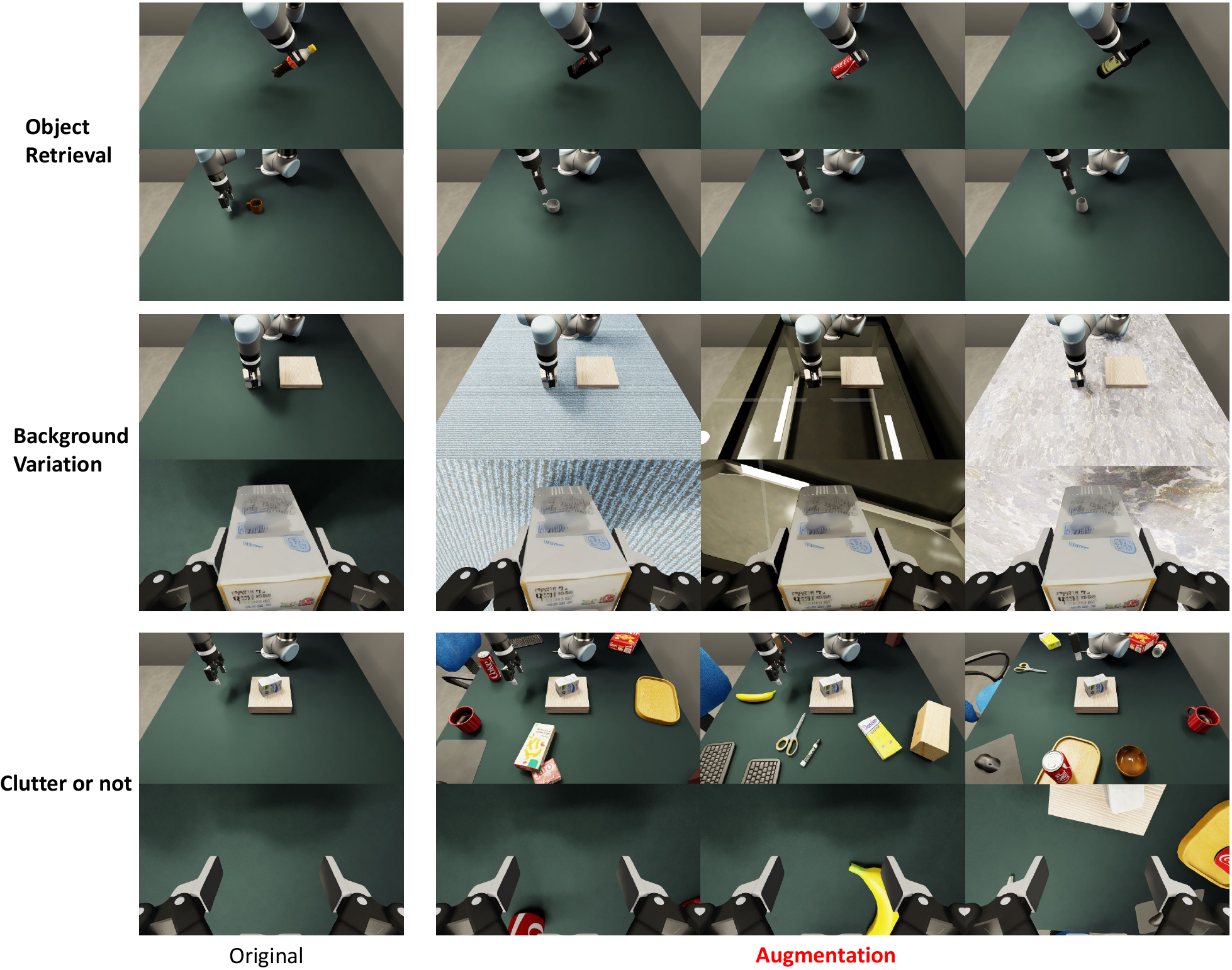}
    \end{overpic}
    \caption{Diverse background texture augmentation  in \nbname.}
    \label{fig:textureaug}
\end{figure*}

%\paragraph{Background Variation}
As shown in ~\cref{fig:textureaug}, we apply scene-level visual randomization to diversify the pixel distribution while keeping task dynamics and contact semantics unchanged: \textit{i}) workspace and background appearance randomization (e.g., tabletop, backsplash) via texture and normal-map swaps, and adjustment of basic PBR parameters (albedo/roughness); \textit{ii}) illumination randomized using parametric light sources with variations in spatial placement, intensity, color, color temperature, and emission radius, enabling a broad range of plausible lighting conditions; \textit{iii}) clutter regime ranging from empty scenes to heavy distractors, with randomly sampled object positions and orientations placed collision-free outside the robot’s swept volume via rejection sampling; \textit{iv}) mild camera intrinsics/extrinsics jitter consistent with prior calibration to emulate plausible view changes.

\subsection{Object Retrival}
\begin{figure*}[!h]
    \centering
    \begin{overpic}[width=0.95\linewidth]{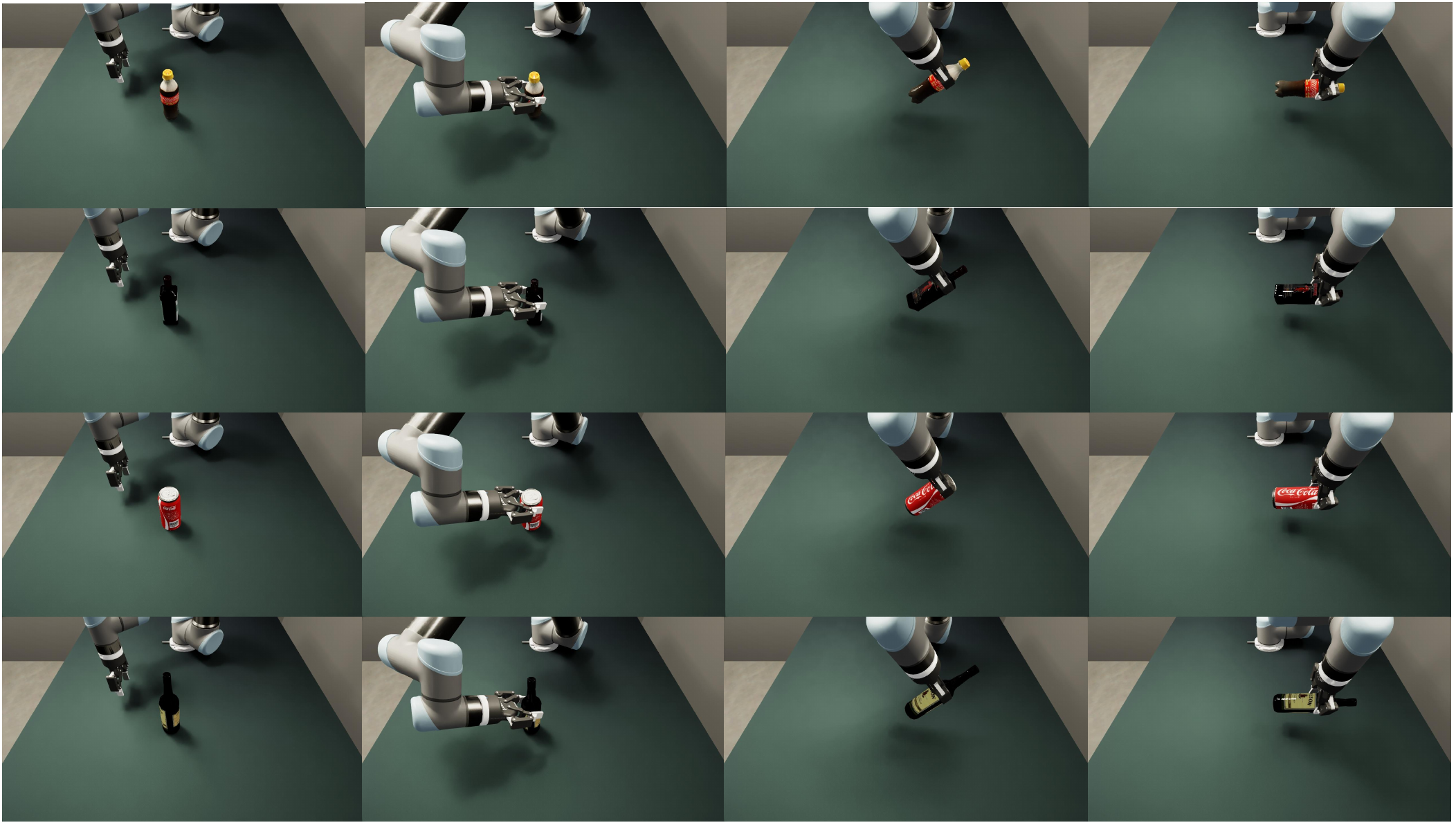}
    \end{overpic}
    \caption{Object Retrieval augmentation}
    \label{fig:append_objaug1}
\end{figure*}

\begin{figure*}[!h]
    \centering
    \begin{overpic}[width=0.95\linewidth]{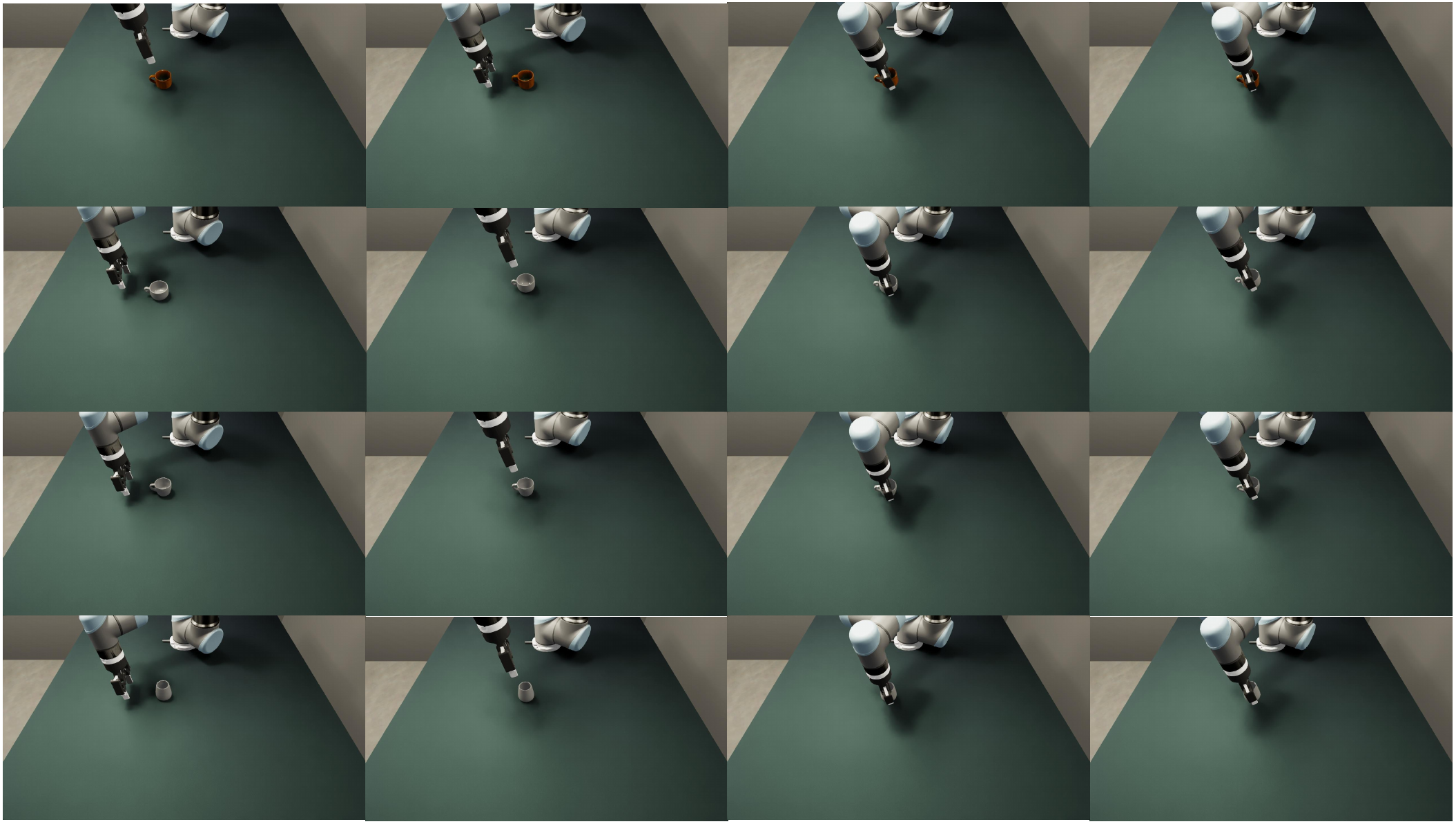}
    \end{overpic}
    \caption{Object Retrieval augmentation}
    \label{fig:append_objaug2}
\end{figure*}

\begin{figure*}[!h]
    \centering
    \begin{overpic}[width=0.95\linewidth]{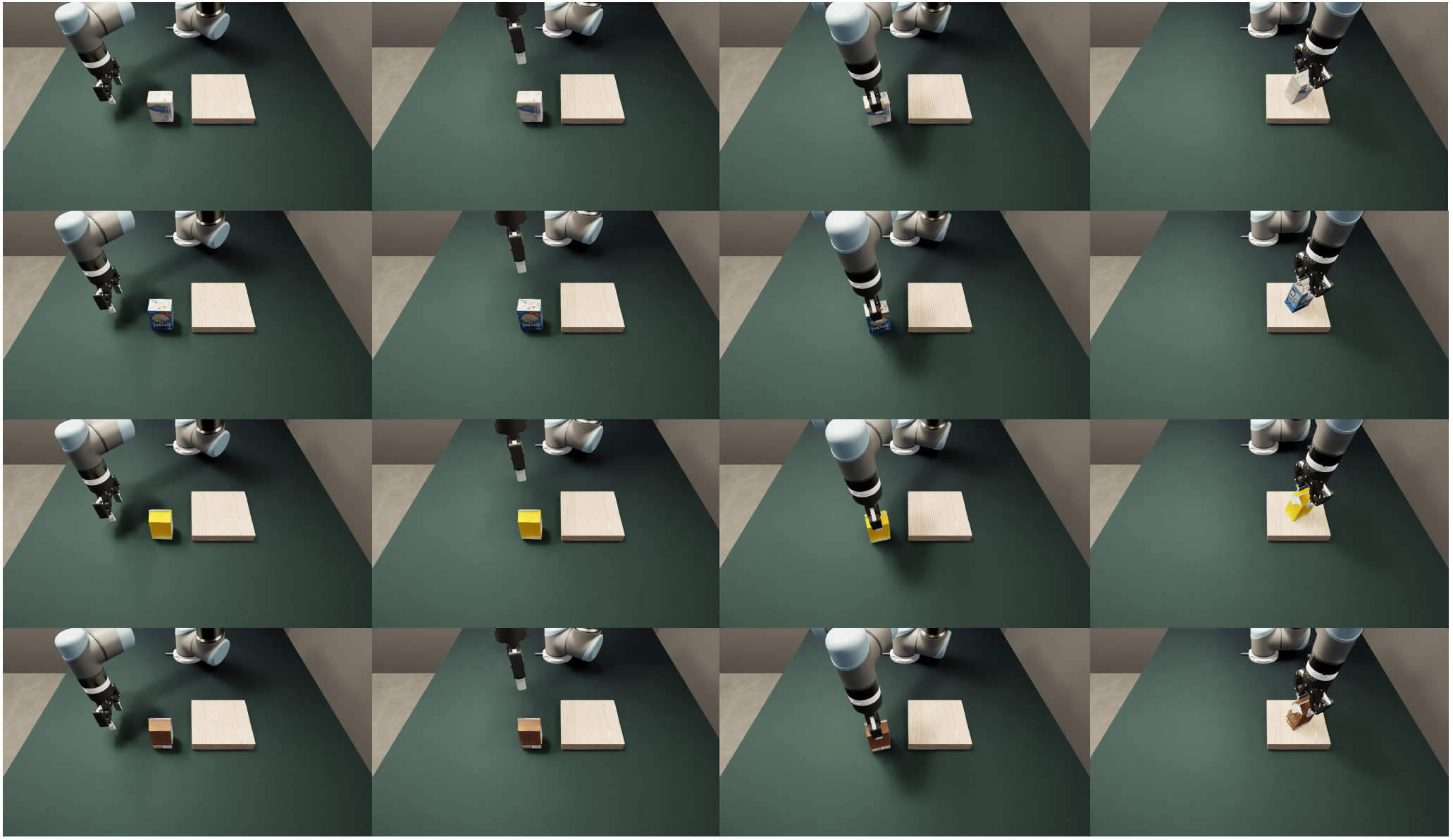}
    \end{overpic}
    \caption{Object Retrieval augmentation}
    \label{fig:append_objaug3}
\end{figure*}

    Our object-retrieval augmentation strategy successfully enables the transfer of manipulation skills to novel objects in simulation. By replacing the original object with a retrieved counterpart that shares high geometric and semantic similarity, and initializing it in the same canonical pose, the robot can reliably execute the same action trajectory. Visually confirmed in Fig.~\ref{fig:append_objaug1}, Fig.~\ref{fig:append_objaug2}, and Fig.~\ref{fig:append_objaug3} for tasks including "pour water", "tip tea cup", and "place box".

\subsection{Hand Mirror}
% \paragraph{Hand-Mirror Augmentation (Left–Right Chirality)}
\textbf{Motivation.} Many daily manipulations are left–right symmetric up to a sagittal-plane reflection; mirroring increases trajectory diversity without changing task semantics.

\textbf{Operator.} Let the sagittal reflection be \(S=\mathrm{diag}(-1,1,1)\). For positions,
\(p'(t)=S\,p(t)\). A pure reflection is improper for orientations, so we compose a \(\pi\)-rotation about the \(y\)-axis to recover a proper rotation:
\begin{equation}
\label{eq:mirror}
R'(t) \;=\; S\,R(t)\,S \cdot R_y(\pi), \qquad \det\!\big(R'(t)\big)=+1.
\end{equation}
We mirror \emph{both} hand and object about the same plane so that
\(
T'_{\mathrm{rel}}(t)=T_h'(t)^{-1}T_o'(t)=T_{\mathrm{rel}}(t)
\),
preserving contact frames and approach vectors. Gripper chirality and finger-axis signs are flipped consistently.

\textbf{Safeguards.} We exclude actions whose handedness encodes semantics (e.g., threaded fasteners), detected via a non-zero screw component about the task \(z\)-axis exceeding \(\tau_{\mathrm{screw}}\). Mirrored rollouts must pass the replay check on a reference arm before inclusion.


\end{document}